%% file: main.tex
\pdfoutput=1
\documentclass[11pt]{article}

\usepackage{latex/acl}

\usepackage{times}
\usepackage{latexsym}
\usepackage{multirow}
\usepackage{graphicx}
\usepackage{booktabs}
\usepackage{amsmath}
\usepackage{amssymb}
\usepackage{amsfonts}
\usepackage{xspace}
\usepackage{color}

\usepackage{enumitem}
\setitemize[1]{itemsep=0pt,partopsep=0pt,parsep=\parskip,topsep=5pt,leftmargin=10pt}

\usepackage[T1]{fontenc}
\usepackage[utf8]{inputenc}

\usepackage{microtype}

\usepackage{inconsolata}

\newcommand\comfact{$\mathcal{C}{om}\mathcal{F}{act}$}
\newcommand\personachat{\textsc{Persona-Chat}}

\newcommand\atomicTT{\textsc{Atomic}$_{20}^{20}$}

\newcommand\comet{\textsc{Comet}}

\newcommand\eg{\textit{e.g.}}
\newcommand\ie{\textit{i.e.}}

\newcommand\wrt{\textit{w.r.t.}}
\newcommand\diffucomet{\textsc{DiffuCOMET}}

\title{\textsc{DiffuCOMET}: Contextual Commonsense Knowledge Diffusion}
\author{\textbf{Silin Gao$^{1}$, Mete Ismayilzada$^{1,3}$, Mengjie Zhao$^{2}$, Hiromi Wakaki$^{2}$} \\
\textbf{Yuki Mitsufuji$^{2}$, Antoine Bosselut$^{1\dagger}$} \\
$^1$NLP Lab, IC, EPFL, Switzerland, $^2$Sony Group Corporation, Tokyo, Japan \\
$^3$Idiap Research Institute, Switzerland \\
{\tt $^1$\{silin.gao,mahammad.ismayilzada,antoine.bosselut\}@epfl.ch} \\
{\tt $^2$\{mengjie.zhao,hiromi.wakaki,yuhki.mitsufuji\}@sony.com} % \\
}

\begin{document}
\maketitle
\renewcommand{\thefootnote}{\fnsymbol{footnote}}
% \footnotetext[1]{Equal contribution.}
\footnotetext[2]{Corresponding author.}
\renewcommand{\thefootnote}{\arabic{footnote}}
\begin{abstract}
%Understanding natural language narratives requires inferring implicit commonsense knowledge that underlies what is explicitly stated in a narrative.
% However, producing contextually-relevant and diverse commonsense inferences remains challenging for current knowledge models that are trained to generate inference. %, and there is a lack of dedicated evaluation metrics.
% non-contextual knowledge completion, surface-form factual knowledge extraction from contexts, or inefficient knowledge linking from limited KBs to contexts.

Inferring contextually-relevant and diverse commonsense to understand narratives remains challenging for knowledge models.
% -- pretrained language models finetuned to generate commonsense knowledge. 
In this work, we develop a series of knowledge models, \diffucomet{}, 
%to better address the task of contextual commonsense knowledge generation.
% Our model series, \diffucomet{}, 
that leverage diffusion to learn to reconstruct the implicit semantic connections between narrative contexts and relevant commonsense knowledge.
Across multiple diffusion steps, our method progressively refines a representation of commonsense facts that is anchored to a narrative, producing contextually-relevant and diverse commonsense inferences for an input context. 
% , which outperforms various existing knowledge models on both social and factual commonsense generation benchmarks.
To evaluate \diffucomet, we introduce new metrics for commonsense inference that more closely measure knowledge diversity and contextual relevance. Our results on two different benchmarks, \comfact{} and WebNLG+, show that knowledge generated by \diffucomet{} achieves a better trade-off between commonsense diversity, contextual relevance and alignment to known gold references, compared to baseline knowledge models.\footnote{We release our code to the community at \url{https://github.com/Silin159/DiffuCOMET}}
%, which also contains more novel facts beyond the coverage of existing annotations.
% In zero-shot narrative adaptation experiments, \diffucomet{} shows better generalization ability to new narrative contexts.
% Moreover, our modeling method generalizes well to factual knowledge generation outside the commonsense domain.
% Besides, our models achieve better performances in human evaluation, which demonstrates the positive correlation between our benchmark metrics and human judgements. % \silin{to be tested}
\end{abstract}

\input{sections/intro}

\input{sections/background}

\input{sections/method}

\input{sections/metrics}
\input{sections/settings}

\input{sections/results}

\input{sections/related_work}

\input{sections/conclusion}

\section*{Limitations}
We outline the following limitations in our work.
First, narrative samples in our training datasets, \ie{}, \comfact{} \citep{gao2022comfact} and WebNLG+ 2020 \citep{ferreira20202020}, have short context windows (five sentences at maximum). Therefore, our knowledge models trained on these datasets may have limited inference capacities if applied to longer narratives that involve richer commonsense grounding.
Moreover, our models are trained on solely English corpora, and may need additional resources to be adapted to other languages or multilingual settings.
Finally, our diffusion modeling method is tested on an encoder-decoder model structure, \ie{}, BART \citep{lewis2020bart}, with maximum model size 406M (BART-large). We leave the feasibility of our method on other model structures, \eg{} decoder-only GPT \citep{radford2019language}, and larger model scales, to future work.

\bibliography{main}

\input{sections/appendix}

\end{document}

%% file: sections/intro.tex
% \begin{figure*}[t]
% \centering
% \includegraphics[width=1.0\textwidth]{pics/method_overview.pdf}
% \caption{Overview of our diffusion-based contextual commonsense knowledge generation. Dashed arrows denote the forward process used for constructing gold references at the training phase. Solid arrows denote the reverse process used for generating commonsense knowledge with attention to the narrative context.}
% \label{method_overview}
% \end{figure*}

\begin{figure}[t]
\centering
\includegraphics[width=1.0\columnwidth]{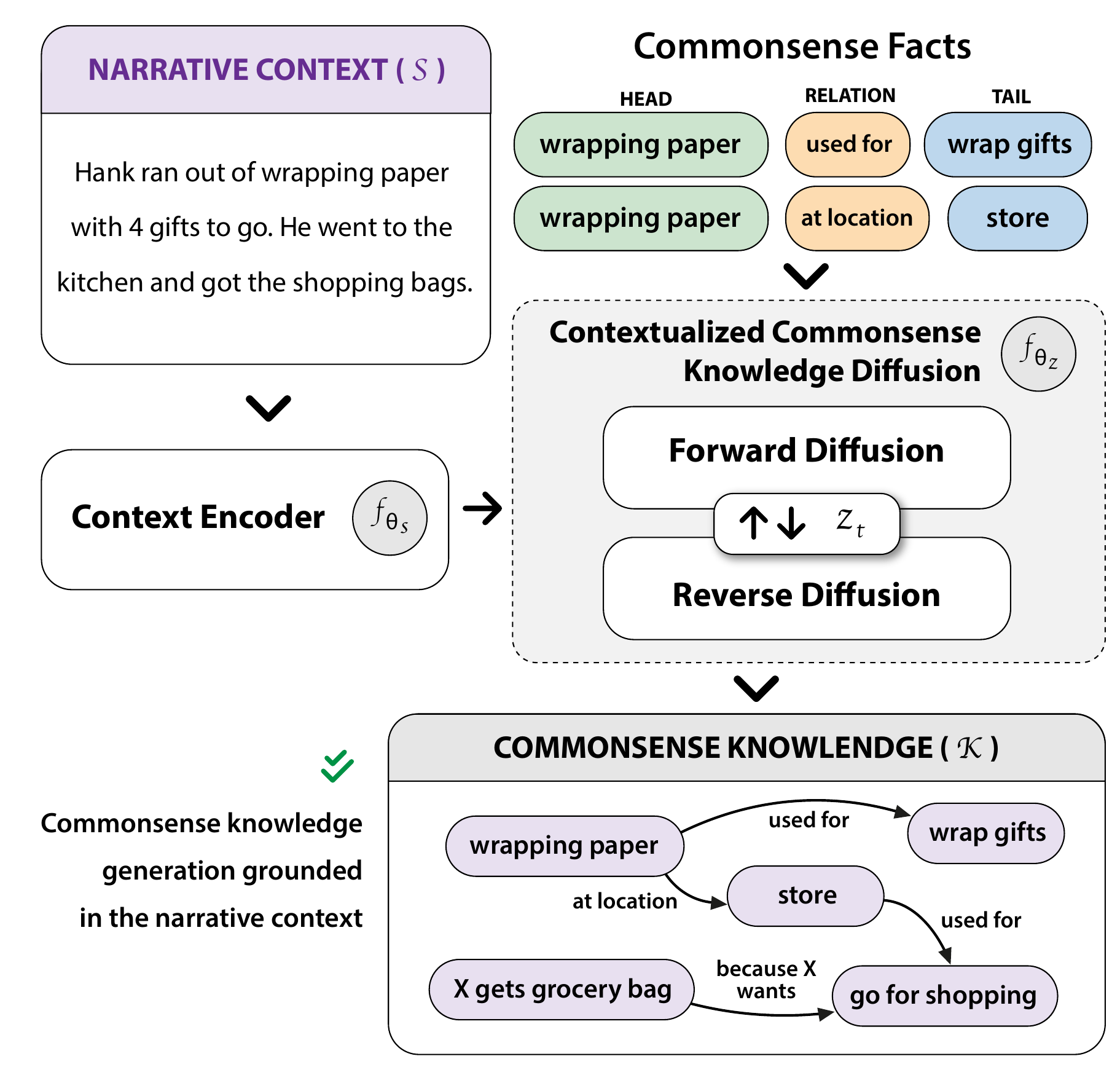}
\caption{Overview of our diffusion-based contextual commonsense knowledge generation.}
\label{method_overview}
\end{figure}

% \begin{figure}[t]
% \centering
% \includegraphics[width=1.0\columnwidth]{pics/method_fact_entity.pdf}
% \caption{Knowledge diffusion based on facts or entities. Dashed arrows denote the forward process used for constructing gold references at the training phase. Solid arrows denote the reverse process used for generating knowledge with attention to the narrative context.}
% \label{method_fact_entity}
% \end{figure}

\section{Introduction}

Identifying the commonsense inferences that underlie narratives, such as stories or dialogues \citep{guan2019story,zhou2022think}, is crucial to understanding those same narratives. For example, to understand why ``Hank ... got the shopping bags'' in the context in Figure~\ref{method_overview}, a model would need to infer that (1) Hank was not finished wrapping gifts, and so (2) would need to buy more wrapping paper. However, comprehensively inferring these diverse, yet implicit, commonsense inferences that are relevant to a context remains a challenging task. % \citep{gao2022comfact}.
Recent methods for identifying contextually-relevant commonsense inferences \citep{bosselut2021dynamic,tu2022misc,peng2022inferring} use knowledge models \citep{bosselut2019comet,west2022symbolic} to generate commonsense facts. While knowledge models have been less brittle than previous retrieval-based methods for commonsense inference, they have two major shortcomings. First, they are trained to verbalize tuples from general commonsense knowledge graphs \citep{sap2019atomic,hwang2021comet}, leading them to produce valid, but often contextually-irrelevant, commonsense inferences when applied out-of-the-box to real narratives. Second, because they are trained using autoregressive training objectives, they subsequently decode high-likelihood, non-diverse sequences that only identify limited collections of commonsense inferences relevant to an input context. %While sampling algorithms can increase the range of generated commonsense inferences produced by knowledge models, they also produce more invalid inferences, and therefore irrelevant inferences.

In this work, we address these challenges of contextual commonsense knowledge generation by developing \textbf{Diffu}sion \citep{ho2020denoising} \textbf{COM}mons\textbf{E}nse \textbf{T}ransformer \citep{bosselut2019comet} models. \diffucomet{} models (shown in Figure~\ref{method_overview}) %learns an alignment between symbolic commonsense knowledge and fact-level (or entity-level) embedding representation, and 
uses diffusion-based decoding to generate relevant knowledge embeddings that are constrained to the narrative context.
Over multiple iterations of constrained diffusion, our models refine a latent representation of the semantic connections between a context and its relevant facts, ensuring that it generates commonsense knowledge that is more contextually relevant to the narrative.
% To integrate discrete knowledge into a continuous diffusion space, \diffucomet{} first learns a mapping between symbolic commonsense knowledge and its fact-level (or entity-level) embedding representations.
% Then, it 
At the same time, by jointly refining multiple fact embeddings during diffusion, \diffucomet{} also generates more diverse inferences than comparable-size autoregressive knowledge models. % that learn using autoregressive training objectives. % greedy or beam search decoding.

We evaluate \diffucomet{} models using traditional NLG metrics (\eg, BLEU; \citealp{papineni2002bleu}) commonly used for evaluating knowledge models. However, these metrics focus on surface form matching to gold references, and fall short of measuring the diversity of commonsense inferences and their semantic relevance to real narrative contexts. Our second contribution is a novel set of metrics that assess the diversity and contextual relevance of knowledge generated by knowledge models. Using both the traditional evaluation metrics and our new suite, we evaluate our models on a commonsense inference linking benchmark \citep{gao2022comfact} that covers both social and physical knowledge, and a second knowledge generation benchmark that involves extracting RDF triplets from language, WebNLG+ \citep{ferreira20202020}.

% For the evaluation of contextual knowledge generation, traditional metrics (\eg, BLEU; \citealp{papineni2002bleu}) for natural language generation (NLG) mainly focus on surface form matching to gold references, which fall short of measuring the diversity of commonsense inferences and their semantic relevance to the context.
% Therefore, we develop a set of new metrics to assess the diversity and contextual relevance of generated knowledge.
% Based on the NLG metrics and our own proposed measures, we evaluate our models on a commonsense construction benchmark \citep{gao2022comfact} that covers both social and physical knowledge.

Our result show that \diffucomet{} models generate knowledge that achieves a better balance of diversity and contextual relevance %and alignment to gold references, 
compared to other knowledge models.
% complex social commonsense \citep{gao2022comfact} and RDF\footnote{Abbreviation of Resource Description Framework} factual commonsense \citep{ferreira20202020} knowledge construction benchmarks.
% On both benchmarks, compared to various autoregressive baseline models, our diffusion knowledge models achieve overall better trade-off between commonsense generation diversity and accuracy.
\diffucomet{} models also more robustly generalize to generate knowledge for out-of-distribution narratives, and are better at producing novel knowledge tuples that are not in their initial training set. % that is beyond the coverage of gold annotations.
%Through zero-shot experiments on narrative contexts sampled from unseen datasets, \diffucomet{} models also demonstrate more robust performances when adapting to out-of-distribution narratives.
Finally, on our second benchmark, WebNLG+, we verify that our diffusion modeling method also generalizes well to a completely new factual knowledge generation task beyond the commonsense domain.

%% file: sections/background.tex
\section{Background: Diffusion Models}
\label{sec:preliminary}
% Diffusion models generate synthetic data based on a pair of forward and backward processes.
% In the forward process, training samples are constructed by gradually adding random noise to input data instances.
% Diffusion models learn the backward process, which gradually de-noises the corrupted samples to recover the original inputs.
% At inference time, synthetic data samples are generated by de-noising pure random noise.
%
% At inference time, synthetic data samples are generated by de-noising random noise through a learned backward process. %that is trained to gradually de-noise corrupted data samples. 
% To learn an alignment between the noise distribution and the output data distribution, the noise distribution is constructed by corrupting input data instances (the same ones that will be de-noised in the backward process) by gradually adding random noise to these instances.
Diffusion models learn to construct synthetic data from random noise.
They use a forward process to gradually corrupt real data samples with additive noise, and learn a reverse process to recover (or de-noise) the corrupted data samples.
Through the de-noising of corrupted data, diffusion models learn to map from a random noise distribution to their target data distribution, which grounds their synthetic data generation.

In this paper, we adopt the DDPM\footnote{\textbf{D}enoising \textbf{D}iffusion \textbf{P}robabilistic \textbf{M}odels} \citep{ho2020denoising} formulation of the forward and reverse diffusion processes.
Specifically, based on a sample $\mathbf{z}_{0}$ from a continuous input data distribution $q(\mathbf{z}_{0})$, the forward process constructs noisy sample $\mathbf{z}_{t}$ over a sequence of time steps $t\in\{1,2,...,T\}$.
In DDPM, $\mathbf{z}_{t}$ is sampled from a Gaussian distribution conditioned on the previous sample $\mathbf{z}_{t-1}$, given by:
\begin{align}
    q(\mathbf{z}_{t}|\mathbf{z}_{t-1})=\mathcal{N}(\mathbf{z}_{t};\sqrt{1-\beta_{t}}\mathbf{z}_{t-1},\beta_{t}\mathbf{I})
    \label{eq:forward}
\end{align}
where $\beta{}_{t}$ is a noise schedule hyperparameter unique to each diffusion step.

In the reverse process, diffusion models learn an inverse distribution $q(\mathbf{z}_{t-1}|\mathbf{z}_{t})$ to de-noise samples created by the forward process.
% To produce more precise predictions of the final de-noised sample $\mathbf{z}_{0}$, 
To more precisely couple the intermediate states of the reverse process with the final de-noised sample $\mathbf{z}_{0}$, 
% Diffusion-LM \citep{li2022diffusion} reformulates the task of predicting $\mathbf{z}_{t-1}$ (based on $\mathbf{z}_{t}$ at time step $t$) as first predicting $\mathbf{z}_{0}$ (from $\mathbf{z}_{t}$) and then reconstructing $\mathbf{z}_{t-1}$ via the forward process defined by Eq. (\ref{eq:forward}).
Diffusion-LM \citep{li2022diffusion} reformulates the task of predicting $\mathbf{z}_{t-1}$ as directly predicting $\mathbf{z}_{0}$ (based on $\mathbf{z}_{t}$), and uses a mean-squared error training loss on the $\mathbf{z}_{0}$ prediction at each time step\footnote{We include more detailed formulation of the reverse diffusion training in Appendix~\ref{appendix:backward_process}.}:
\begin{align}
\mathcal{L}_{z_{0}\mbox{-}mse} = \sum_{t=1}^{T} \mathbb{E} \Vert \mathbf{z}_{0} - f_{\theta}(\mathbf{z}_{t},t) \Vert^{2}
\label{eq:z0-mse-loss}
\end{align}
where $f_{\theta}(\mathbf{z}_{t},t)=\hat{\mathbf{z}}^{t-1}_{0}$ denotes the model's learned prediction of $\mathbf{z}_{0}$ at the reverse stage of step $t$ to $t-1$.
To formulate $\hat{\mathbf{z}}^{t-1}_{0}$ as a refinement of the former reverse stage's output $\hat{\mathbf{z}}^{t}_{0}$, Bit-Diffusion \citep{chen2022analog} improves the model function of predicting $\mathbf{z}_{0}$ with self-conditioning, \ie{}, $\hat{\mathbf{z}}^{t-1}_{0}=f_{\theta}(\hat{\mathbf{z}}^{t}_{0},\mathbf{z}_{t},t)$.
At inference time, the noisy sample at step $t$ is predicted from $\hat{\mathbf{z}}^{t}_{0}$ via the Eq.~(\ref{eq:forward}) forward process, denoted as $\hat{\mathbf{z}}_{t}$ to replace the unknown gold input $\mathbf{z}_{t}$, while the initial input $\mathbf{z}_{T}$ is pure Gaussian noise sampled from $\mathcal{N}(\mathbf{0},\mathbf{I})$.

%% file: sections/method.tex
% \section{Method}

% In this work, we use diffusion methods to construct contextually-relevant commonsense knowledge graphs. In this section, we review diffusion models, and introduce two novel methods for using diffusion for commonsense knowledge generation.

\begin{figure}[t]
\centering
\includegraphics[width=1.0\columnwidth]{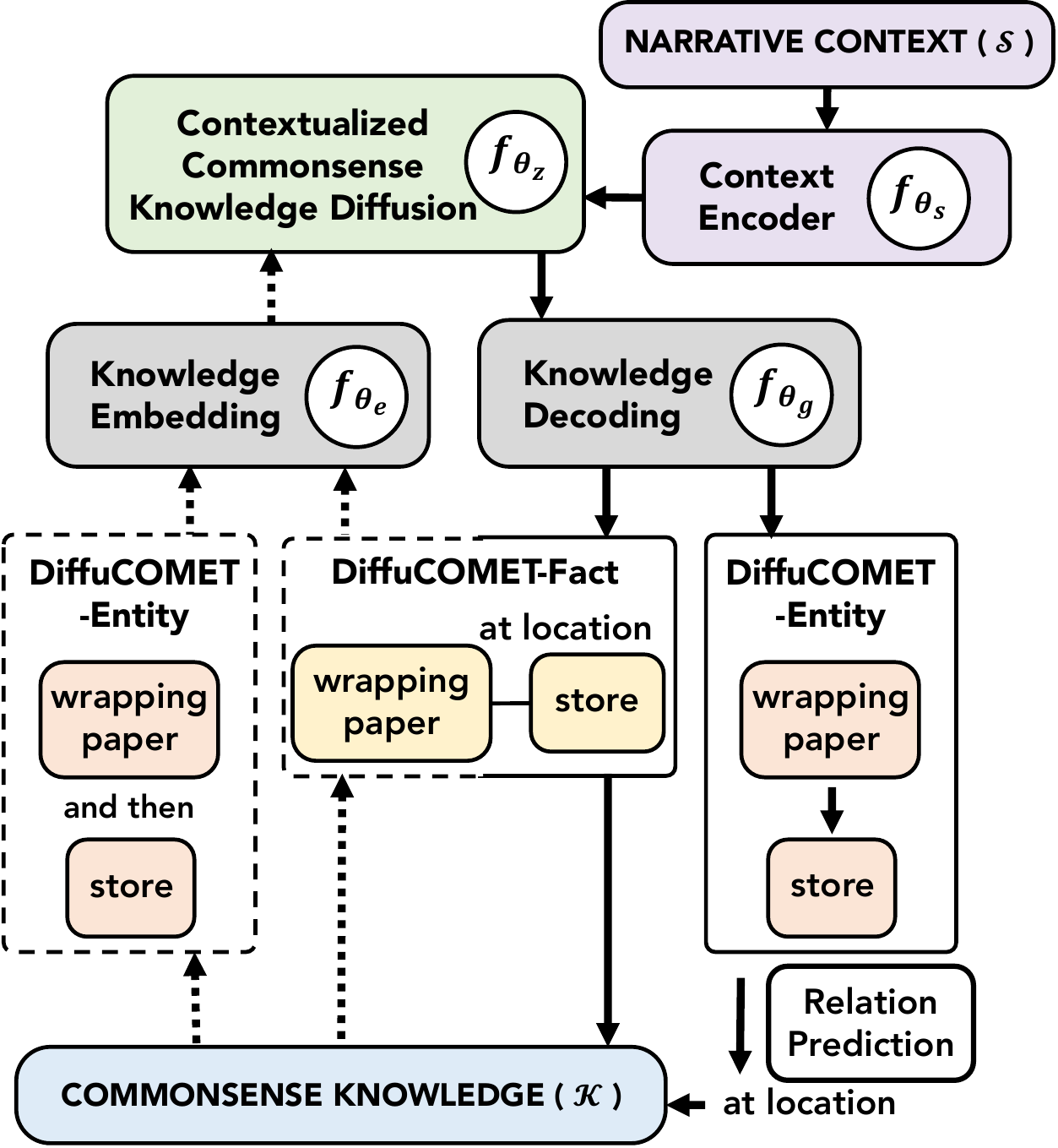}
\caption{Knowledge diffusion based on facts or entities. Dashed arrows denote the forward process used for constructing gold references at the training phase. Solid arrows denote the reverse process used for generating knowledge with attention to the narrative context.}
\label{method_fact_entity}
\end{figure}

\section{Contextual Knowledge Diffusion}
In this section, we first introduce the task of contextual commonsense knowledge generation, and then propose \diffucomet{}, our diffusion approach for this task. The overview of our method is presented in Figure~\ref{method_overview}.

\paragraph{Task Description}
Given a narrative sample $\mathcal{S}$ as context, \eg{}, a story snippet or a dialogue, the model needs to generate commonsense inferences as a set facts $\mathcal{K}=\{k_{1},...,k_n,...,k_{N}\}$, which are relevant for understanding the situation described in the context.
Each fact $k_{n}=(h_{n},r_{n},a_{n})$ is represented as a triple containing a head entity $h_{n}$, a tail (attribute) entity $a_{n}$, and a relation $r_{n}$ connecting them, \eg{}, (\textit{wrapping paper}, \textit{used for}, \textit{wrap gifts}), as shown in Figure~\ref{method_overview}.
We denote the set of unique head entities, relations and tail entities in $\mathcal{K}$ as $\mathcal{H}$, $\mathcal{R}$ and $\mathcal{A}$, respectively.

\paragraph{Contextualization}
We ground knowledge diffusion on the given context $\mathcal{S}$ by using encoder-decoder cross attention, inspired by SeqDiffuSeq \citep{yuan2022seqdiffuseq}.
In particular, we use a BART \citep{lewis2020bart} encoder $f_{\theta_{s}}$ to learn the context encoding that represents $\mathcal{S}$ as hidden state $\mathbf{z}_{\mathcal{S}}$:
\begin{align}
\mathbf{z}_{\mathcal{S}}=f_{\theta_{s}}(\mathcal{S})
\label{eq:context}
\end{align}
Then, a BART decoder $f_{\theta_{z}}$, serving as the diffusion module, learns to predict the de-noised data sample $\mathbf{z}_{0}$.
% with cross-attention to the hidden state of the context.
Given the context hidden state $\mathbf{z}_{\mathcal{S}}$ (via cross-attention to the encoder $f_{\theta_{s}}$), $f_{\theta_{z}}$ makes a prediction of $\mathbf{z}_{0}$ at time step $t$-1 (\ie, $\hat{\mathbf{z}}^{t-1}_{0}$) based on its former prediction $\hat{\mathbf{z}}^{t}_{0}$ and time step $t$'s noisy sample $\mathbf{z}_{t}$:
\begin{align}
\hat{\mathbf{z}}^{t-1}_{0}=f_{\theta_{z}}(\hat{\mathbf{z}}^{t}_{0},\mathbf{z}_{t},t|\mathbf{z}_{\mathcal{S}})
\label{eq:diff_module}
\end{align}
Unlike traditional transformer decoders \citep{vaswani2017attention}, the diffusion module $f_{\theta_{z}}$ applies a bi-directional self-attention to $\hat{\mathbf{z}}^{t}_{0}$ and $\mathbf{z}_{t}$, since all positions of $\hat{\mathbf{z}}^{t-1}_{0}$ are decoded simultaneously, \ie{}, in non-autoregressive manner.\footnote{More implementation details of the diffusion module $f_{\theta_{z}}$ are presented in Appendix~\ref{appendix:diffusion_module}.}

\paragraph{Discrete Knowledge Diffusion}
We consider two formulations for representing discrete knowledge in continuous embedding spaces for diffusion: \textbf{\diffucomet-Fact}, where we learn to reconstruct continuous representations of facts $k_{n}$ using diffusion, and \textbf{\diffucomet-Entity}, where we use separate diffusion processes to reconstruct head $h_n$ and tail $a_n$ representations and then predict the relation between them to complete the fact. We highlight these differences in Figure~\ref{method_fact_entity}.

For diffusion on the fact-level embedding space (\textbf{\diffucomet-Fact}), we first pre-train a BART encoder $f_{\theta_{e}}$ to produce an embedding $\mathbf{e}_{n}$ of each fact $k_{n}$ in the knowledge set $\mathcal{K}$ (with embedding size $d$ same as the hidden state size of BART):
\begin{align}
\mathbf{e}_{n}=f_{\theta_{e}}(k_{n}) \in \mathbb{R}^{d}
\label{eq:embed}
\end{align}
% \abm{what's $f_{\theta_{e}}$}
where we input the concatenation of each fact's head, relation and tail tokens to the encoder $f_{\theta_{e}}$, and take the output hidden state of a start token \textit{<s>} as the embedding of the fact.
The initial input $\mathbf{z}_{0}$ of the forward diffusion process is then sampled from a Gaussian centered on the concatenation of all fact embeddings $\mathbf{e}=[\mathbf{e}_{1};\mathbf{e}_{2};...;\mathbf{e}_{|\mathcal{K}|}] \in \mathbb{R}^{d \times |\mathcal{K}|}$, formulated as $q_{e}(\mathbf{z}_{0}|\mathbf{e})=\mathcal{N}(\mathbf{z}_{0};\mathbf{e},\beta_{0}\mathbf{I})$.

In the reverse process, the diffusion module $f_{\theta_{z}}$ is trained to generate the final output $\hat{\mathbf{z}}^{0}_{0}$ (using time step $1$'s input $\mathbf{z}_{1}$ and $\hat{\mathbf{z}}^{1}_{0}$) as its predicted fact embeddings $\hat{\mathbf{e}}$, \ie{}, $\hat{\mathbf{e}}=\hat{\mathbf{z}}^{0}_{0}=f_{\theta_{z}}(\hat{\mathbf{z}}^{1}_{0},\mathbf{z}_{1},1|\mathbf{z}_{\mathcal{S}})$.
Finally, we pre-train another BART decoder $f_{\theta_{g}}$ to generate the synthetic fact $\hat{k}_{n}$ with conditioned on the diffusion module's predicted $n$-th embedding $\hat{\mathbf{e}}_{n}=\hat{\mathbf{e}}[:][n]$,  ($n=1,2,...,|\mathcal{K}|$)\footnote{At inference time, the maximum value of $n$ (number of generated facts) can be arbitrary depending on the user's choice. In Appendix~\ref{appendix:fact_number}, we introduce how we control the number of facts that our models generate for each context.}:
\begin{align}
\hat{k}_{n}=f_{\theta_{g}}(\hat{\mathbf{e}}_{n})
\label{eq:decode}
\end{align}

\noindent For diffusion on the entity-level embedding space (\textbf{\diffucomet-Entity}), we use a pipeline to generate head entities, tail entities and their relations.
First, to generate head entities, we use a similar process as in \textbf{\diffucomet-Fact}, \ie{}, pre-train a BART encoder to produce a gold embedding of each unique head entity ${h}_{i}\in\mathcal{H}$ (for training the diffusion module), and then pre-train a BART decoder to generate synthetic head entities $\hat{h}_{i}$ from the diffusion module's predicted embeddings.
%, formulated as Eq. (\ref{eq:embed}) with input fact $k_{n}$ replaced by head entity $h_{i}$ ($i=1,2,...|\mathcal{H}|$).
%, formulated as Eq. (\ref{eq:decode}) with fact generation $\hat{k}_{n}$ replaced by head generation $\hat{h}_{i}$.
Each predicted head entity $\hat{h}_{i}$ is then appended to the context (\ie{}, $\mathcal{S}$ in Eq. \ref{eq:context}), expanding the context to $\mathcal{S}_{i}=[\mathcal{S},\hat{h}_{i}]$.
A second diffusion module predicts embeddings of synthetic tail entities $\hat{a}_{j}$ related to $\mathcal{S}_{i}$ (trained using gold embeddings of tail entities $a_{j}\in\mathcal{A}$ that possess relations $r_{ij}\in\mathcal{R}$ to the gold head ${h}_{i}$).
A final BART model predicts the relation $\hat{r}_{ij}$ between each pair of generated head and tail entities, grounded on the context.
% \begin{align}
% \hat{r}_{ij}=f_{\theta_{r}}(\hat{h}_{i}, \hat{a}_{j}, \mathcal{S})
% \end{align}

\paragraph{Embedding Module Training}
We pretrain the embedding modules ($f_{\theta_{e}}$, $f_{\theta_{g}}$), which focus on modeling generic knowledge representations independent to the context, before the diffusion modules ($f_{\theta_{s}}$, $f_{\theta_{z}}$), which learn the specific mapping from the context to its relevant knowledge. When training the diffusion modules, we freeze the pre-trained embedding modules.

To pretrain the fact (or entity) embedding modules, we minimize the decoder's negative log-likelihood of re-constructing facts $k$ (or entity $h$ or $a$) in the full set of knowledge $\mathcal{K}_{full}$ involved in the whole narrative dataset (or domain), based on its embedding given by encoder $f_{\theta_{e}}$:
\begin{align}
\mathcal{L}_{\theta_{e},\theta_{g}} = - \log p_{\theta_{g}}(k|f_{\theta_{e}}(k))
\label{eq:embed_loss}
\end{align}

\paragraph{Diffusion Module Training} We optimize a dual loss to train the diffusion modules. First, we consider the mean-square error loss of the diffusion module's de-noised sample prediction $\hat{\mathbf{z}}^{t}_{0}$ at each time step $t$, compared to the reference sample $\mathbf{z}_{0}$ (for $t>0$) and gold embeddings $\mathbf{e}$ (for $t=0$):
\begin{align}
\mathcal{L}^{mse}_{\theta_{s},\theta_{z}} = \mathbb{E} \Vert \mathbf{e} - \hat{\mathbf{z}}^{0}_{0} \Vert^{2} + \sum_{t=1}^{T-1} \mathbb{E} \Vert \mathbf{z}_{0} - \hat{\mathbf{z}}^{t}_{0} \Vert^{2}
\label{eq:mse}
\end{align}
We also use an anchor loss \citep{gao2022difformer} to supervise the final fact (or entity) generation.
For each time step $t$, we minimize the negative log-likelihood of the embedding module decoder (with frozen parameters $\theta_{g}$) generating each fact $k_{n}$ in knowledge set $\mathcal{K}$, based on the diffusion module's predicted de-noised sample $\hat{\mathbf{z}}^{t}_{0}$:
\begin{align}
\mathcal{L}^{gen}_{\theta_{s},\theta_{z}} = \sum_{t=0}^{T-1} \sum_{n=1}^{|\mathcal{K}|} - \log p_{\theta_{g}}(k_{n}|\hat{\mathbf{z}}^{t}_{0}[:][n])
\label{eq:anchor}
\end{align}
where $\hat{\mathbf{z}}^{t}_{0}[:][n]$ is the predicted de-noised representation of $k_{n}$.
The final loss is $\mathcal{L}_{\theta_{s},\theta_{z}} = \mathcal{L}^{mse}_{\theta_{s},\theta_{z}} + \gamma\mathcal{L}^{gen}_{\theta_{s},\theta_{z}}$, where $\gamma$ is a tunable hyperparameter.

\paragraph{Inference} At inference time, the reverse diffusion process is initialized with noise sampled from the Gaussian distribution $\mathcal{N}(\mathbf{0},\mathbf{I})$, while the embedding module encoder $f_{\theta_{e}}$, which provides gold diffusion references for training, is not used.

%% file: sections/metrics.tex
\begin{figure*}[t]
\centering
\includegraphics[width=1.0\textwidth]{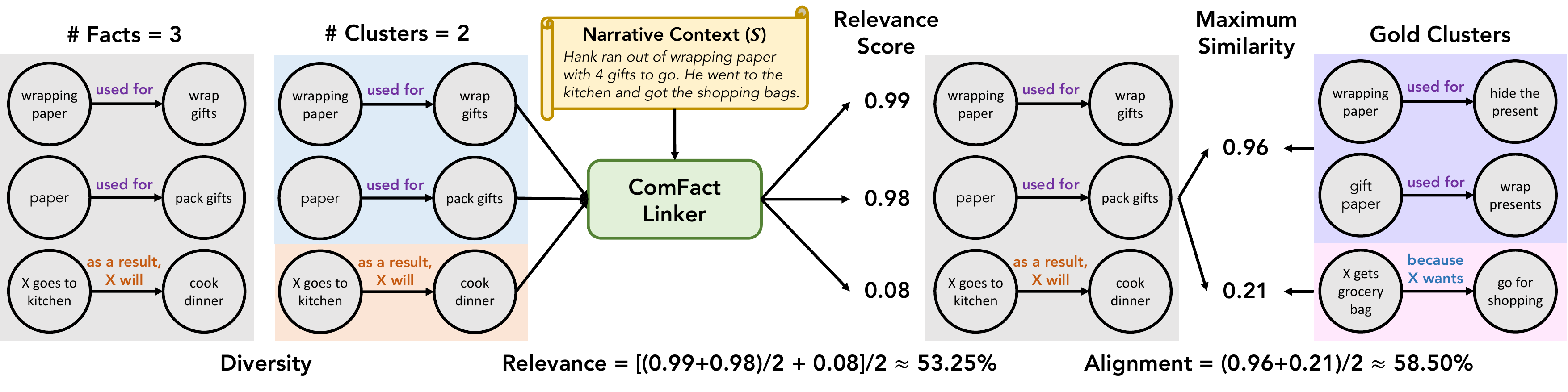}
\caption{Illustration of clustering-based evaluation metrics for contextual commonsense knowledge generation.}
\label{metrics}
\end{figure*}

\section{Evaluation}
\label{sec:cluster_metrics}
Prior work in commonsense knowledge generation \citep{hwang2021comet,da2021analyzing} evaluated knowledge models using traditional NLG metrics (\eg, BLEU; \citealp{papineni2002bleu}) in controlled studies with KGs, where the inputs to the models were head entities and relations and the knowledge model produced tail attributes.
In practice, however, knowledge models are used to generate implicit commonsense inferences for natural language contexts \citep{ismayilzada2023kogito}, requiring generated inferences to be relevant to a more complex input than a basic KG head entity, and necessitating diverse generated inferences that comprehensively augment the context. However, traditional NLG metrics fall short of measuring these important dimensions because they measure surface form overlap between model outputs and references, which rewards generating facts with similar or duplicated semantics, limiting diversity. 
% which fail to measure several important dimensions of commonsense knowledge generation. 
% First, they only measure surface form alignment between model outputs and references, ignoring implicit semantic similarity. Second, they reward generating facts with similar or duplicated semantics, ignoring diversity. 
% Furthermore, prior works evaluated in controlled studies using KGs, where the inputs to the models were head entities and relations and the knowledge model produced tail attributes.
% In practice, knowledge models are often used to generate relevant inferences for natural language contexts \citep{ismayilzada2023kogito}, requiring generated knowledge to both cover a diverse array of possible commonsense inferences, and ensure these generated inference remain relevant to the context. 

Motivated by these shortcomings, we propose novel evaluation metrics that assess the diversity and contextual relevance of generated knowledge. First, to eliminate the effect of knowledge repetition in generations, we cluster similar facts and treat each fact cluster (instead of each single fact) as a unit piece of knowledge.
In particular, we use the DBSCAN \citep{ester1996density} algorithm to group gold facts $\mathcal{K}=\{{k_{1},k_{2},...,k_{N}}\}$ and generated facts $\hat{\mathcal{K}}=\{\hat{k}_{1},\hat{k}_{2},...,\hat{k}_{\hat{N}}\}$ into clusters $\mathcal{C}=\{c_{1},c_{2},...,c_{M}\}$ and $\hat{\mathcal{C}}=\{\hat{c}_{1},\hat{c}_{2},...,\hat{c}_{\hat{M}}\}$, respectively.
We test two methods for measuring the similarity of facts for clustering: word-level edit distance \citep{levenshtein1966binary}, which measures the difference of two facts' surface-form tokens, and Euclidean distance of Sentence-BERT \citep{reimers2019sentence} embeddings, which measures the semantic difference of two facts.
Based on these clusters, we develop three metrics to measure the diversity of generated facts, their contextual relevance, and their alignment to gold references, as shown in Figure~\ref{metrics}. 
% We describe our metrics below:

\paragraph{Diversity.} To measure the diversity of generated facts (\ie, amount of distinctive knowledge being generated), we count the number of fact clusters (\textbf{\#~Clusters}), \ie{}, $\hat{M}$ (or $M$ for gold references). 
We also report the number of facts (\textbf{\#~Facts}), \ie{}, $\hat{N}$ (or $N$ for gold references), to compare the number of fact clusters to the number of generated facts produced by the models.

% \abm{could this Relevance measure be biased ? if your approach is trained on ComFact and the baselines aren't, would the model be more precise at predicting your outputs} 

\paragraph{Relevance.} We measure the relevance of the fact clusters to the narrative context using a fact linker\footnote{Fact linking models predict the relevance of knowledge tuples to textual passages \citep{gao2022comfact}} trained on the \comfact{} dataset \citep{gao2022comfact} %(based on DeBERTa; \citealp{he2020deberta})
that scores the relevance of each fact $\hat{k}_{n}$ to the context $\mathcal{S}$, denoted as $rel(\hat{k}_{n},\mathcal{S})\in[0,1]$.
% \footnote{We assume that in each round of evaluation, all models target to generate the same domain of commonsense knowledge, so the \comfact{} linker is not biased on any specific model's output knowledge type.}
The relevance score of a fact cluster $\hat{c}_{m}$ is defined as the average relevance score of its facts, \ie{}, $\sum_{\hat{k}_{n}\in\hat{c}_{m}} rel(\hat{k}_{n},\mathcal{S})/|\hat{c}_{m}|$.
Finally, we measure the average relevance over all fact clusters in $\hat{\mathcal{C}}$:
% \footnote{We present a human evaluation in Appendix~\ref{sec:full_results} to verify that the \comfact{} linker's measurement of relevance is aligned with human judgements.}
\begin{equation}
    rel(\hat{\mathcal{C}},\mathcal{S})=\frac{1}{\hat{M}} \sum_{\hat{c}_{m}\in\hat{\mathcal{C}}} \frac{1}{|\hat{c}_{m}|} \sum_{\hat{k}_{n}\in\hat{c}_{m}} rel(\hat{k}_{n},\mathcal{S})
\end{equation}

\noindent We note that \textbf{Relevance} can be viewed as a  \textit{precision} measure for generated facts, which tends to decrease as more facts are generated because irrelevant facts are more likely to be generated.

\paragraph{Alignment} measures the average similarity of generated facts to gold fact clusters.
Specifically, we define a function $sim(\hat{k}_{i},k_{j})\in[0,1]$ to measure the pairwise similarity between a generated fact and a gold reference (using similar distance functions to define clusters above\footnote{Further details on exact definitions are in Appendix~\ref{appendix:similarity_alignment}.}).
% \ab{this formula doesn't make sense}
%, so we cut-off the negative similarity scores to $0$
Using this function, we measure the maximum pairwise similarity of generated facts to references in each gold cluster $c_{m}$$\in\mathcal{C}$, which serves as the alignment score to the gold cluster.
Finally, we average the alignment scores of generated facts to all gold clusters:
\begin{equation}
    sim(\hat{\mathcal{K}},\mathcal{C})=\frac{1}{M} \sum_{c_{m}\in\mathcal{C}} \max_{\substack{\hat{k}_{i}\in\hat{\mathcal{K}}, \\ k_{j}\in c_{m}}}sim(\hat{k}_{i},k_{j})
\end{equation}
%$sim(\hat{\mathcal{K}},\mathcal{C})=\frac{1}{M} \sum_{c_{m}\in\mathcal{C}} \max_{\hat{k}_{i}\in\hat{\mathcal{K}},k_{j}\in c_{m}}sim(\hat{k}_{i},k_{j})$.

\noindent We note that \textbf{Alignment} can be viewed as the generated facts' \textit{recall} of gold fact clusters, which tends to increase as more facts are generated because more facts will be aligned to gold clusters. % \ie{}, more likely to hit gold references.
%While \textbf{Relevance} can be viewed as a \textit{precision} measure of generated fact clusters, which by contrast is prone to decrease as more facts are generated, \ie{}, harder to discover more relevant facts.
Given this trade-off between Relevance and Alignment, we also present the harmonic mean of Relevance and Alignment as an overall evaluation of the two dimensions, denoted as \textbf{RA-F1}.

%% file: sections/settings.tex
\begin{table*}[t]
\centering
\resizebox{1.0\textwidth}{!}{
\smallskip\begin{tabular}{lcccccccc}
\hline
% \multirow{2}*{\textbf{Model}} & \multirow{2}*{\textbf{\# Facts}}  & \multicolumn{5}{c}{\textbf{Clustering \wrt{} Word-Level Edit Distance}} & \multicolumn{5}{c}{\textbf{Clustering \wrt{} Embedding Euclidean Distance}}\\
%             \cmidrule(lr){3-7} \cmidrule(lr){8-12}
\textbf{Model} & \textbf{\# Facts} & \textbf{\# Clusters} & \textbf{Relevance} & \textbf{Alignment} & \textbf{RA-F1} & \textbf{BLEU} & \textbf{METEOR} & \textbf{ROUGE-L} \\
\hline
Greedy-\comet{} & 1.96 & 1.19 & 61.42 & 50.64 & 55.51 & \textbf{18.01} & \textbf{52.32} & \textbf{54.96} \\
Sampling-\comet{} & 15.00 & \textbf{8.39} & 56.19 & \textbf{77.97} & 65.31 & 12.69 & 44.43 & 45.58 \\
Beam-BART & 15.00 & 4.60 & 64.35 & 71.35 & 67.67 & 13.11 & 47.70 & 46.35 \\
Beam-\comet{} & 15.00 & 5.09 & 65.03 & 71.64 & 68.18 & 16.97 & 47.39 & 47.19 \\
Grapher & 5.08 & 2.60 & \textbf{68.29} & 40.58 & 50.91 & 1.40 & 23.96 & 27.21 \\
\midrule
\diffucomet-Fact   & 12.88 & 5.24 & 65.64 & 71.65 & \underline{68.51} & 15.98 & \underline{50.06} & \underline{51.44} \\
\diffucomet-Entity & 12.89 & \underline{5.67} & \underline{66.39} & \underline{74.38} & \textbf{70.16} & \underline{17.01} & 47.61 & 48.40 \\
\midrule
Gold  & 10.55 & 5.64 & 80.90 & - & - & - & - & - \\
\hline
\end{tabular}
}
\caption{Evaluation results on the ROCStories portion of \comfact{}. Both \diffucomet{} models presented are developed based on BART-large. Models with suffix ``-\comet{}'' and ``-BART'' are fine-tuned on \comet{}-BART and BART-large. Presented results of our proposed metrics are based on fact clustering \wrt{} embedding Euclidean distance. Best and second-best results (excluding Gold references) are \textbf{bolded} and \underline{underlined}, respectively.}
\label{tab:main_results_roc}
\end{table*}

% \ab{I would just pick one of these distance functions for the main result (maybe have a table for the other one in the appendix? Then put Table 2 with the raw metrics in the main table. I'm not sure there's much benefit to putting both BART-base and BART-large results in the main paper.}

\section{Experimental Settings}
\label{sec:settings}
\paragraph{Datasets}First, we evaluate our approach on the \comfact{} \citep{gao2022comfact} benchmark, where models need to generate \atomicTT{} \citep{hwang2021comet} social commonsense facts that are relevant to narrative contexts sampled from four diverse corpora: \personachat{} \cite{zhang2018personalizing}, MuTual \citep{cui2020mutual}, ROCStories \cite{mostafazadeh2016corpus} and CMU Movie Summary \citep{bamman2013learning}.
We only use training data from the ROCStories portion of \comfact{}, to enable the evaluation of zero-shot generalization on the other three partitions of the dataset.
Our fact embedding module is pretrained on the full \atomicTT{} knowledge base, which contains $\sim972K$ commonsense facts after preprocessing.\footnote{More data preprocessing details are in Appendix~\ref{appendix:data_preprocess}.} We also evaluate our approach in a conceptually different setting, the WebNLG+ 2020 \citep{ferreira20202020} dataset, which consists of RDF \citep{ora1999resource} facts sampled from the DBpedia \citep{lehmann2015dbpedia} knowledge base, with corresponding natural language texts verbalizations.
The task is to generate the sampled RDF facts given their verbalizations.
% We develop our models based on the $\sim35K$ WebNLG training texts and their linked RDF facts, following the same data pre-processing used in the Grapher \citep{melnyk2022knowledge} baseline model.
We use $\sim$$13$k facts from the training data to pretrain the fact embedding module.

% PersonaExt \citep{zhu2023paed} persona attribute extraction

\paragraph{Baselines}
We train \diffucomet{} using BART-base and BART-large as pretrained models, and compare with three baselines developed on the same backbones: a) a \textbf{Greedy} baseline that is trained to autoregressively generate the concatenation of all relevant facts,\footnote{Facts are concatenated by a special token \textit{<fsep>}.} b) a \textbf{Sampling} baseline that uses nucleus sampling \citep{holtzman2019curious} to generate multiple individual facts in parallel, and c) a Diverse \textbf{Beam} search baseline that uses diverse beam search to generate multiple inferences in parallel.
We also compare our models trained using BART-large to baselines developed on models of similar scale: d) the aforementioned greedy decoding, sampling and beam search baselines trained from \textbf{\comet}-BART \citep{hwang2021comet}, a BART-large model further pre-trained on \atomicTT{} for commonsense knowledge completion, and e) \textbf{Grapher} \citep{melnyk2022knowledge}, which trains a T5-large \citep{raffel2020exploring} model to generate entities (nodes) related to the context, followed by a MLP classifier to predict the relations (edges) between entities.

\paragraph{Metrics} We evaluate these methods on our clustering-based metrics described in Section~\ref{sec:cluster_metrics}.
As the clustering algorithm (\ie{}, DBSCAN) used in our metrics has an adjustable clustering granularity controlled by a distance threshold, we consider a range of distance thresholds and take the average of evaluation results across all thresholds in the range, allowing us to avoid biasing our metrics to a specific distance threshold.\footnote{We include more details of our clustering threshold selection in Appendix~\ref{appendix:clustering_threshold}.}
% For the choice of threshold range, we consider an equal interval at where the number of gold fact clusters significantly varies from near the maximum (\ie{}, \# Facts) to near the minimum (\ie{}, $1$).
For \comfact{}, we also test on the metrics from \citealp{hwang2021comet} for evaluating commonsense knowledge generation, including \textbf{BLEU} \citep{papineni2002bleu}, \textbf{METEOR} \citep{banerjee2005meteor} and \textbf{ROUGE-L} \citep{lin2004rouge}. %, as well as a common measure for diversity, \textbf{Distinct-4} \citep{li2016diversity}, which measures the ratio of distinct 4-grams to all 4-grams appearing in generated facts.
For evaluation on WebNLG+ 2020, we also report the official metrics for this dataset's challenge \citep{ferreira20202020}, which construct optimal pairings between predicted facts and gold references, and then compute precision, recall, and F1 scores based on the surface-form matching of paired facts. We denote these WebNLG metrics as \textbf{Web-Prec.}, \textbf{Web-Rec.} and \textbf{Web-F1}.\footnote{More details of WebNLG metrics are in Appendix~\ref{appendix:webnlg_metrics}.}

%% file: sections/results.tex
\section{Results and Analysis}

% \begin{table*}[t]
% \centering
% \resizebox{1.0\textwidth}{!}{
% \smallskip\begin{tabular}{lccccccccc}
% \toprule
% \multirow{2}*{\textbf{Model}} & \multicolumn{3}{c}{\textbf{PersonaChat}} & \multicolumn{3}{c}{\textbf{MuTual}} & \multicolumn{3}{c}{\textbf{MovieSummaries}}\\
%             \cmidrule(lr){2-4} \cmidrule(lr){5-7} \cmidrule(lr){8-10}
%  & \textbf{\# Facts} & \textbf{\# Clusters}  & \textbf{RA-F1} & \textbf{\# Facts} & \textbf{\# Clusters}  & \textbf{RA-F1} & \textbf{\# Facts} & \textbf{\# Clusters}  & \textbf{RA-F1} \\
% \hline
% Beam-BART & 15.00 & 4.04 & 58.27 & 15.00 & 4.45 & 64.77 & 15.00 & 3.70 & \underline{46.15} \\ 
% Beam-\comet{} & 15.00 & \underline{4.27} & 59.04 & 15.00 & 4.75 & 64.82 & 15.00 & \underline{4.04} & 45.88 \\
% Grapher & 4.53 & 1.57 & 41.24 & 4.50 & 1.78 & 54.31 & 5.34 & 1.50 & 41.97 \\
% \midrule
% \diffucomet-Fact  & 10.82 & 3.89 & \underline{59.68} & 10.46 & \underline{4.80} & \underline{66.04} & 8.29 & 2.89 & 43.51 \\
% \diffucomet-Entity & 12.06 & \textbf{4.48} & \textbf{60.50} & 11.85 & % \textbf{5.39} & \textbf{67.32} & 13.50 & \textbf{5.84} & \textbf{50.42} \\
% \midrule
% Gold  & 8.60 & 4.28 & - & 10.80 & 5.79 & - & 9.00 & 4.81 & - \\
% \bottomrule
% \end{tabular}
% }
% \caption{Zero-shot evaluation results on the PersonaChat, MuTual and MovieSummaries portions of \comfact{} benchmark. Notations are the same as Table~\ref{tab:main_results_roc}, where best and second-best results are in bold and underlined.}
% \label{tab:results_zero_shot_comfact}
% \end{table*}

% \subsection{\comfact{} Benchmark}

Table~\ref{tab:main_results_roc} shows evaluation results on the ROCStories portion of the \comfact{} benchmark for our \diffucomet{} models developed based on BART-large.\footnote{Presented results of our metrics are based on fact clustering \wrt{} embedding Euclidean distance. Results based on word-level edit distance are included in Appendix~\ref{appendix:roc_stories}, and promote the same conclusions.}
% We present our \diffucomet{} models developed based on BART-large, and compare to baselines with similar-scale model backbones, as described in Section~\ref{sec:settings}.
% We mainly present the baseline models trained on \comet-BART, which achieve overall better performances than baselines trained on BART-large, as illustrated by the comparison between Beam-BART and Beam-\comet.\footnote{Full results of all baseline models are in Appendix~\ref{sec:full_results}.}
% Note that we include the comparisons of models with BART-base backbones in Appendix~\ref{sec:full_results}, where our models outperform baselines with a larger gap, \ie{}, $\sim15\%$ absolute RA-F1 improvement in average.
%
On our new cluster-based metrics, \diffucomet{} models demonstrate a better balance between diversity and accuracy in contextual knowledge generation.
Specifically, \diffucomet{} models achieve Relevance and Alignment scores that are both comparable to the best baseline results, contributing to their higher overall RA-F1 measures, while also producing a larger number of distinct knowledge clusters. 
% By contrast, Greedy, Sampling and Grapher baseline models present imbalanced trade-off between knowledge diversity and accuracy, which significantly sacrifice one or two dimensions of quality among \#~Clusters, Relevance and Alignment.
% Compared to Beam models that achieve better diversity-accuracy balance than other baselines, \diffucomet{} models achieve consistently higher scores on all evaluation dimensions.
By contrast, the Greedy, Sampling and Grapher baselines significantly sacrifice one or two dimensions of diversity and quality \wrt{} \#~Clusters, Relevance and Alignment. Beam baselines consistently underperform \diffucomet{} on cluster metrics.
% All above results demonstrate diffusion models' superior feasibility in text-to-knowledge generation.

For evaluation on the traditional NLG metrics, we find that \diffucomet{} models score higher overall than most baseline models on metrics that check the alignment with gold references, \ie{}, BLEU, METEOR and ROUGE-L, except for the Greedy decoding baseline, whose higher scores are artificially high because it generates very little knowledge, \ie{}, only $\sim$$2$ facts per context.
% This again verifies the outstanding accuracy of diffusion models' knowledge generation.
%While \diffucomet{} performs worse on Distinct-4 than a few baselines, this metric is also biased toward systems that produce few generations, \eg{}, Greedy and Grapher. 
We also include further comparisons of models with BART-\textit{base} backbones in Appendix~\ref{appendix:roc_stories}, where our models outperform baselines by a larger gap, \ie{}, $\sim$$15\%$ absolute RA-F1 improvement on average.

We also test \diffucomet{}'s ability to generalize to out-of-domain contexts using the other portions of \comfact{} with contexts sampled from PersonaChat, MuTual and MovieSummaries. We report generalization results to the above three portions in Appendix Tables~\ref{tab:results_cluster_full_persona}-\ref{tab:results_nlg_full_movie}, and observe similar results where \diffucomet{}-Entity outperforms baselines by $\sim$$5\%$ RA-F1 and produces $\sim$$20\%$ more knowledge clusters.

% From zero-shot table at top of doc
% RA-F1
% ((50.42 + 67.32 + 60.5) - (46.15 + 59.04 + 64.82)) / (46.15 + 59.04 + 64.82)
% # Clusters
% ((4.48+5.39 + 5.84) - (4.27 + 4.75 + 4.04)) / (4.27 + 4.75 + 4.04)

\begin{table}[t]
\small
\centering
% \resizebox{1.0\columnwidth}{!}{
\smallskip\begin{tabular}{lcc}
\toprule
% \small
\textbf{Model} & \textbf{Validity} & \textbf{Relevance} \\
\midrule
Sampling-\comet{} & 49.45 & 30.20 \\
Beam-\comet{} & \textbf{74.80} & 42.81 \\
% Grapher & 34.86 & 32.27 \\
\midrule
\diffucomet-Fact & 70.00 & \underline{48.27} \\
\diffucomet-Entity & \underline{74.15} & \textbf{54.18} \\
\midrule
Gold & 94.79 & 82.04 \\
\bottomrule
\end{tabular}
% }
\caption{Human evaluation results.}% Notations are same as Table~\ref{tab:main_results_roc}.}
\label{tab:human_eval}
\end{table}

\begin{table}[t]
\centering
\resizebox{1.0\columnwidth}{!}{
\smallskip\begin{tabular}{lcc}
\toprule
\textbf{Model} & \textbf{\# Novel Facts} & \textbf{\# Novel Clusters} \\
\midrule
Sampling-\comet{} & 0.26 & 0.19 \\
Beam-\comet{} & 0.27 & 0.17 \\
\midrule
\diffucomet-Fact & \textbf{0.30} & \underline{0.20} \\
\diffucomet-Entity & \textbf{0.30} & \textbf{0.24} \\
\bottomrule
\end{tabular}
}
\caption{Novelty of generated knowledge.}%novel knowledge detection on \comfact{} ROCStories. Notations are same as Table~\ref{tab:main_results_roc}.}
\label{tab:novelty}
\end{table}

The results of our automatic evaluation are also supported by our human evaluation.
We hire Amazon Mechanical Turk workers\footnote{Details on workers and their payment are in Appendix~\ref{appendix:human_evaluation}} to evaluate the validity and contextual relevance of models' generated knowledge on the ROCStories portion of \comfact{}.
Specifically, given a narrative context and a list of commonsense facts that a model generates about the context, we ask three workers to independently judge whether each fact is valid and relevant\footnote{\textit{invalid} facts are automatically labeled \textit{irrelevant}} to the context, and take their majority vote as the assessment.
% We label a fact as relevant if it passes the verification of two or all three workers.
%Finally, we report the number of valid and relevant facts that the model generates for each context. 
In Table~\ref{tab:human_eval}, we see that \diffucomet{} models produce \textit{valid} facts at about the same rate as the best baseline, but produce facts that are far more relevant to the narrative context. %Interestingly, humans tend to prefer the outputs of \diffucomet-Entity by a slight margin. %our human evaluation results in Appendix~\ref{appendix:human_evaluation}, and find that the model rankings rated by human workers and our \textbf{Relevance} metric are aligned with each other, verifying the reliability of using neural fact linkers as relevance scorers.

% % \paragraph{Context Generalization}
% We further test \diffucomet{}'s ability to generalize to out-of-domain contexts using the other three portions of \comfact{} benchmark, where contexts are sampled from PersonaChat, MuTual and MovieSummaries, respectively.
% We report zero-shot generalization results to the above three portions in Appendix Tables~\ref{tab:results_cluster_full_persona}-\ref{tab:results_nlg_full_movie}, where we observe that \diffucomet{} models keep outperforming baselines on the contexts of PersonaChat and MuTual.
% % In Table~\ref{tab:results_zero_shot_comfact}, we observe that both \diffucomet{} models generalize well to the contexts of PersonaChat and MuTual, whose generated knowledge possesses comparable diversity (\ie{}, \#~Clusters) and better accuracy (\ie{}, RA-F1) than the strongest baseline model Beam-\comet{}.
% On the more challenging MovieSummaries-style contexts that involve relatively longer and more complex narratives, \diffucomet{}-Entity shows better generalization ability than \diffucomet{}-Fact, achieving larger points of improvements over baselines.
% More interestingly, we find that \diffucomet{}-Entity achieves larger points of improvements over baselines on the more challenging MovieSummaries-style contexts, while \diffucomet{}-Fact struggles to outperform the strongest baseline Beam-\comet{}, showing that entity-level diffusion is more robust to the shift of narrative contexts, likely due to the more fine-grained multi-step learning of context-to-knowledge mapping.

\begin{figure}[t]
\centering
\includegraphics[width=1.0\columnwidth]{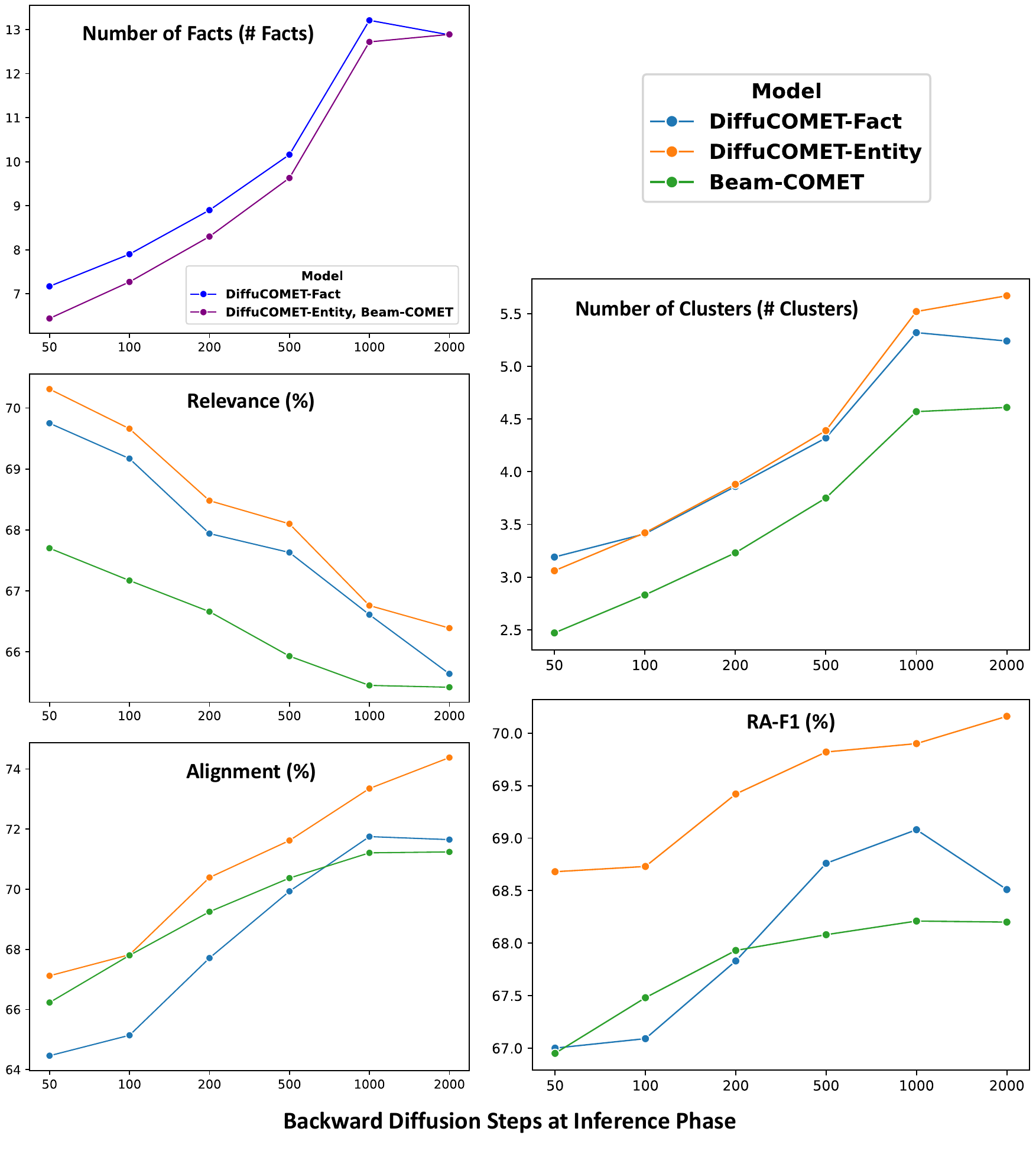}
\caption{\diffucomet{} performance at different diffusion steps during inference. Both \diffucomet{}-Fact and \diffucomet{}-Entity are developed based on BART-large and tested on the ROCStories portion of \comfact{}. Beam-\comet{} performance is shown as a baseline, with the number of decoded facts set to match \diffucomet{}-Entity at each diffusion step.}
\label{diffusion_steps}
\end{figure}

\paragraph{Novelty}
\diffucomet{} models also produce more novel commonsense inferences. A historical advantage of knowledge models (\eg, \comet{}) was their ability to generate knowledge beyond the graphs they used for pretraining \citep{bosselut2019comet}, making them powerful tools to generate commonsense knowledge for unseen narratives. 
% Since knowledge generation models generally have the potential to create new inferences beyond their training knowledge, we also study whether \diffucomet{} can construct novel commonsense that is not covered by gold references.
To test the novelty of generated commonsense knowledge from \diffucomet{}, we develop a heuristic method that identifies knowledge as \textit{novel} if its maximum pair-wise (Sentence-BERT embedding) cosine similarity to \comfact{} gold references is lower than $0.45$.
However, as this cut-off would likely cause invalid and irrelevant facts to be considered novel, we only include facts whose relevance score is higher than $0.97$.\footnote{Thresholds are tuned by a manual check of 100 sampled results to ensure a decent cutoff of novel and relevant facts.}
In Table~\ref{tab:novelty}, we see that \diffucomet{} models produce more novel facts and clusters compared to baselines.\footnote{We conduct analysis on some examples of novel facts in a case study in Appendix~\ref{appendix:case_study}.}
% whose generated numbers of facts (\# Facts) are similar to our models.
% We show a comparison of detected novel facts generated by Beam-\comet{} and \diffucomet{}-Entity in Table~\ref{tab:examples}, where we find that our model constructs more diverse novel knowledge related to physical entities (\eg{}, cap) and social interactions (\eg{}, connections between family and vacation), while by comparison, Beam-\comet{}'s novelty is mainly restricted to simple physical knowledge.\footnote{We include more analysis of models' generated types of knowledge in Appendix~\ref{sec:full_results}.}

\paragraph{Diffusion Steps} To investigate how \diffucomet's multiple rounds of knowledge representation refinement through the diffusion process affect the quality of generated knowledge, we record the performance of our \diffucomet{} models as a function of diffusion steps conducted during inference.
Figure~\ref{diffusion_steps} shows how \diffucomet{}'s performance varies when knowledge is generated at earlier time steps. 

We find that \diffucomet{} models gradually produce more facts and more diverse facts (\ie{}, \#~Clusters) as the number of diffusion steps increase, indicating that the multiple rounds of diffusion produce a more separable representation capable of representing more facts.
% across various number of generated facts (\#~Facts).
While the greater number of facts leads to a slight drop in contextual relevance across the generated facts, a greater corresponding increase in alignment to the gold clusters (as observed by the increase in Alignment and RA-F1) also emerges. %, we verify the trade-off where more facts being generated typically leads to lower precision of contextual relevance but higher recall of gold references (\ie{}, alignment).
% This demonstrates the necessity of our RA-F1 metric.
% , which balances the measure of relevance and alignment.
On RA-F1, \diffucomet{}-Fact surpasses Beam-\comet{}\footnote{To make the comparison intuitive, for each test context, we dynamically set the beam size of Beam-\comet{} to the number of facts generated by \diffucomet{}-Entity.} as the diffusion steps increase to larger than $200$, and \diffucomet{}-Entity consistently scores higher and continues benefiting from further diffusion, even after $1000$ diffusion steps.
These results shows that multi-step refinement of facts via diffusion effectively improves contextual knowledge generation.
%, and also implies that entity-level diffusion possesses better potential to refine ground knowledge of narratives.
% Interestingly, we detect that \diffucomet{}-Fact reaches the best relevance-alignment trade-off (\ie{}, highest RA-F1) around $1000$ diffusion steps, while \diffucomet{}-Entity continuously benefits from more diffusion steps after $1000$.
% This implies that decomposed pipeline entity diffusion possesses better potential to refine ground knowledge of narratives.

% \ab{can we run an experiment where take the diffused output at different steps $t$ and explore the content of the graph ?}

\begin{table}[t]
\centering
\resizebox{1.0\columnwidth}{!}{
\smallskip\begin{tabular}{lccc}
\toprule
% \multirow{2}*{\textbf{Model}} & \multirow{2}*{\textbf{\# Facts}}  & \multicolumn{5}{c}{\textbf{Clustering \wrt{} Word-Level Edit Distance}} & \multicolumn{5}{c}{\textbf{Clustering \wrt{} Embedding Euclidean Distance}}\\
%             \cmidrule(lr){3-7} \cmidrule(lr){8-12}
\textbf{Model} & \textbf{Web-Prec.} & \textbf{Web-Rec.} & \textbf{Web-F1} \\
\midrule
% Greedy-\comet{} & 52.30 & 54.59 & 53.37 \\
% Sampling-\comet{} & 73.77 & 76.67 & 75.15 \\
Beam-BART & 73.36 & 76.27 & 74.75 \\
% Beam-\comet{} & 73.80 & 76.61 & 74.85 \\
Grapher & 71.20 & 73.00 & 71.90 \\
\midrule
\diffucomet-Fact & \underline{76.30} & \underline{78.07} & \underline{77.19} \\
\diffucomet-Entity & \textbf{80.68} & \textbf{82.89} & \textbf{81.74} \\
\bottomrule
\end{tabular}
}
\caption{Results on \textbf{WebNLG+ 2020}. Official metrics used for the benchmark challenge are presented.}% Notations are same as Table~\ref{tab:main_results_roc}.}
\label{tab:main_results_webnlg}
\end{table}

\subsection{WebNLG+ 2020 Benchmark}
Finally, to test whether our method generalizes outside the domain of generating commonsense inferences, we present our evaluation results on the WebNLG+ 2020 dataset in Table~\ref{tab:main_results_webnlg}. \diffucomet{} models achieve better performances on the WebNLG factual knowledge generation task, verified by the official metrics of the benchmark.\footnote{We also include the evaluation results on traditional NLG and our proposed clustering-based metrics in Appendix~\ref{appendix:webnlg_results}.}
This results suggests that our diffusion approach to knowledge graph construction could be adapted to other knowledge generation tasks. %, which also implies that \diffucomet{} models have great potential to generalize to new domains of knowledge.

%% file: sections/related_work.tex
\section{Related Work}
\paragraph{Commonsense Knowledge Grounding}
% Commonsense knowledge often serves as the basis of knowledge-intensive NLP tasks, \eg{}, question answering \citep{feng2020scalable,yasunaga2021qa,zhang2022greaselm,yasunaga2022deep}, dialogue modeling \citep{zhou2018commonsense,wu2020diverse,zhou2022think}, story generation \citep{guan2019story,ji2020language}, and multi-modality \citep{zhang-etal-2022-visual,wu2023towards}.
% To augment NLP systems with commonsense knowledge, all above works typically use retrieval methods based on heuristics or embedding similarity to link relevant facts from commonsense knowledge graphs \citep{liu2004conceptnet,speer2017conceptnet,sap2019atomic,hwang2021comet}.
To augment NLP systems with commonsense knowledge, various systems for question answering \citep{zhang2022greaselm,yasunaga2021qa,yasunaga2022deep} and narrative generation \citep{ji2020language,zhou2022think} use retrieval methods based on heuristics to link relevant facts from commonsense knowledge graphs \citep{speer2017conceptnet,sap2019atomic,gao2023peacok}.
However, these systems typically have low precision when adapted to more general and complex commonsense linking \citep{hwang2021comet,jiang2021m}.
\citealp{gao2022comfact} developed commonsense fact linking to improve retrieval precision, but this requires inefficiently traversing all candidate facts to check their contextual relevance.

% Due to above limitations of retrieval-based knowledge grounding, previous works also consider using generative methods to identify contextually-relevant commonsense knowledge.
% One line of research \citep{bosselut2021dynamic,peng2022inferring,peng2022guiding,tu2022misc} uses knowledge models \citep{bosselut2019comet,da2021analyzing,west2022symbolic} to generate tail inferences from narrative statements. However, these outputs do not form complete knowledge triples, and often produce irrelevant facts as the knowledge models are pre-trained for context-free knowledge graph completion.
Due to above limitations of retrieval-based knowledge grounding, one line of research \citep{bosselut2021dynamic,tu2022misc} uses knowledge models \citep{bosselut2019comet,west2022symbolic} to generate tail inferences from narrative statements.
However, these methods often produce irrelevant facts as the knowledge models are pre-trained for context-free knowledge graph completion.
Finally, developing new knowledge models to learn contextual commonsense generation turns out to be a promising track of research, while current works are limited to simple physical \citep{zhou2022think} or RDF-style factual \citep{melnyk2022knowledge} knowledge.
We build new models to address contextual commonsense generation in a more general scope.

\paragraph{Diffusion Models}
% In the advanced research of computer vision, diffusion models \citep{sohl2015deep,song2019generative,ho2020denoising} achieve outstanding performances in the task of text-to-image generation \citep{rombach2022high,kim2022diffusionclip,bao2023one}.
% Inspired by this success, researchers also apply the power of diffusion models to various NLP applications, \eg{}, controllable review and story generation \citep{li2022diffusion}, sequence-to-sequence generation for dialogue, text rewriting and machine translation \citep{gong2022diffuseq,yuan2022seqdiffuseq}.
% There are also considerable works \citep{han2022ssd,strudel2022self,chen2022analog,gao2022difformer,lin2022genie} on developing more advanced methods to improve the text generation with diffusion.
Considerable recent works \citep{gao2022difformer,lin2022genie,han2024transfer} have developed methods to improve text generation with diffusion models \citep{sohl2015deep,song2019generative,ho2020denoising}.
However, the potential of diffusion models in text-to-knowledge generation is still under-explored.
In this paper, we introduce diffusion models for the task of contextual knowledge generation.

%% file: sections/conclusion.tex
\section{Conclusion}
In this work, we leverage the power of diffusion models for contextual commonsense knowledge generation, and formulate novel metrics to highlight important dimensions of diversity and contextual relevance for this task.
Our diffusion knowledge models, \diffucomet{}, outperform various autoregressive knowledge models, producing more diverse, novel, and contextually-relevant commonsense knowledge, and achieving better out-of-distribution performance.
% In novel knowledge detection and domain adaptation experiments, our models also generalize better to produce novel inferences that are relevant for out-of-distribution narrative contexts. 
Finally, our analysis reveals how \diffucomet{} refines implicit knowledge representations over the course of the diffusion process to produce more relevant and diverse inferences, hinting at our method's potential benefit in other text-to-graph generation tasks.
% which are positively aligned with human evaluation results.
% These superior knowledge generation abilities of diffusion models demonstrate their great potential to benefit downstream knowledge-intensive NLP applications.

%% file: sections/appendix.tex
\appendix

\section{Backward Diffusion Process}
\label{appendix:backward_process}
Inverting from the forward diffusion process formulated as Eq.(\ref{eq:forward}), the backward diffusion process follows a Gaussian posterior distribution $q(\mathbf{z}_{t-1}|\mathbf{z}_{t},\mathbf{z}_{0})$:
\begin{align}
\begin{split}
    q(\mathbf{z}_{t-1}|\mathbf{z}_{t},\mathbf{z}_{0})&=\mathcal{N}(\mathbf{z}_{t-1};\widetilde{\mu}(\mathbf{z}_{t},\mathbf{z}_{0}),\widetilde{\beta_{t}}\mathbf{I}) \\
    \widetilde{\mu}(\mathbf{z}_{t},\mathbf{z}_{0})&=\frac{\sqrt{\overline{\alpha}_{t-1}}\beta_{t}}{1-\overline{\alpha}_{t}}\mathbf{z}_{0} +\frac{\sqrt{\alpha_{t}}(1-\overline{\alpha}_{t-1})}{1-\overline{\alpha}_{t}}\mathbf{z}_{t} \\
    \widetilde{\beta_{t}}&=\frac{1-\overline{\alpha}_{t-1}}{1-\overline{\alpha}_{t}}\beta_{t}
\end{split}
\end{align}
where $\alpha_{t}=1-\beta_{t}$ and $\overline{\alpha}_{t}=\prod_{i=1}^{t}\alpha_{i}$ are weight hyperparameters of the posterior Gaussian defined by the noise schedule $\beta_{t}$.
The posterior formulation indicates that only the mean $\widetilde{\mu}$ of $\mathbf{z}_{t-1}$ is correlated to the condition $\mathbf{z}_{t}$ and $\mathbf{z}_{0}$.
So the training loss for diffusion models, derived from the KL-divergence between gold and learned posterior distributions, is typically defined as a mean-squared error loss on the posterior Gaussian mean:
\begin{align}
\mathcal{L}_{mse} = \sum_{t=1}^{T} \mathbb{E} \Vert \widetilde{\mu}(\mathbf{z}_{t},\mathbf{z}_{0}) - \mu_{\theta}(\mathbf{z}_{t},t) \Vert^{2} 
\end{align}
where model (with parameter $\theta$) learns the function $\mu_{\theta}(\mathbf{z}_{t},t)$ to predict the mean of $\mathbf{z}_{t-1}$.
Diffusion-LM \citep{li2022diffusion} further re-weights the mean-squared error as Eq.(\ref{eq:z0-mse-loss}) to enforce direct prediction of $\mathbf{z}_{0}$ in every loss term, which is shown to be more efficient at tuning the model to precisely predict the final de-noised sample.

\section{Model Implementation Details}
\subsection{Diffusion Module}
\label{appendix:diffusion_module}
To conduct the diffusion process defined by Eq.(\ref{eq:diff_module}) using Transformers \citep{vaswani2017attention}, $\mathbf{z}_{t}$ and $\hat{\mathbf{z}}^{t}_{0}$ are first concatenated at the hidden-state dimension and projected by a MLP layer to form their joint representation.
The positional encoding layer of Transformers is applied to the time step $t$ (same for every position of self-attention), whose output time step embedding is added to the joint representation of $\mathbf{z}_{t}$ and $\hat{\mathbf{z}}^{t}_{0}$.
The decoder $f_{\theta_{z}}$ takes the joint representation (with time step embedding added) as its bi-directional self-attention input, to ground its decoding of refined $\mathbf{z}_{0}$ prediction $\hat{\mathbf{z}}^{t-1}_{0}$.

\subsection{Number of Generated Facts}
\label{appendix:fact_number}
To enable our diffusion module ($f_{\theta_{z}}$) to control the number of facts (or entities) generated for each context, we also pre-train our fact (or entity) embedding module ($f_{\theta_{e}}$ and $f_{\theta_{g}}$) to learn the representation of a special token $k_{end}:=$\textit{<eok>}, by adding it as a special fact (or entity) to the pre-training data, which indicates the end of a knowledge set.
During the training of diffusion module, $k_{end}$ is appended to the end of knowledge set $\mathcal{K}$, whose embedding and decoding also contributes to the training loss.
At inference phase, we post-process our model's generations to keep only the facts that are at positions before $k_{end}$.

\subsection{Noise Schedule}
\label{appendix:noise_schedule}
For the noise schedule hyperparameter of diffusion process, we adopt the \textit{sqrt} initialization \citep{li2022diffusion} to set $\overline{\alpha}_{t}=1-\sqrt{t/T+s}$, where $s=1e^{-4}$ that sets the initial variance of noise ($\beta_{0}$) to be $0.01$.
Based on that, we follow SeqDiffuSeq \citep{yuan2022seqdiffuseq} to implement an adaptive noise schedule, which dynamically adjusts $\overline{\alpha}_{t}$ for each sample position $n$ ($n=1,2,...|\mathcal{K}|$) of the knowledge set $\mathcal{K}$ (the adjusted $\overline{\alpha}_{t}$ for position $n$ is denoted as $\overline{\alpha}^{n}_{t}$), according to the diffusion mean square error (MSE) loss $\mathcal{L}^{mse}_{\theta_{s},\theta_{z}}$ defined in Eq.~(\ref{eq:mse}).
Specifically, for an adaptive noise schedule update, we first record the MSE loss at each time $t$ and position $n$ as:
\begin{align}
\mathcal{L}^{n}_{t}=\mathbb{E} \Vert \mathbf{z}_{0}[:][n] - \hat{\mathbf{z}}^{t}_{0}[:][n] \Vert^{2}
\end{align}
Then we use a linear interpolation function to update the adjusted noise schedule, formulated as: \begin{align}
F^{n}_{t}(\mathcal{L})=\frac{\overline{\alpha}^{n}_{t}-\overline{\alpha}^{n}_{t-1}}{\mathcal{L}^{n}_{t}-\mathcal{L}^{n}_{t-1}}(\mathcal{L}-\mathcal{L}^{n}_{t-1})+\overline{\alpha}^{n}_{t-1}
\end{align}
where new loss value $\mathcal{L}^{n,new}_{t}$ is re-arranged across time step $t$ with equal interval between $\min_{t}(\mathcal{L}^{n}_{t})$ and $\max_{t}(\mathcal{L}^{n}_{t})$, which is finally given to the update function to get $\overline{\alpha}^{n,new}_{t}=F^{n}_{t}(\mathcal{L}^{n,new}_{t})$.
The noise schedule is adjusted every 2000 training steps.

\subsection{Model Training}
\label{appendix:model_training}
For the loss weight hyperparameter $\gamma$ used to combine mean-square error and anchor losses defined by Eq.~(\ref{eq:mse}) and (\ref{eq:anchor}), we use $\gamma=1$ for training our \diffucomet{} models based on BART-base, while $\gamma=0.01$ for training our models with BART-large backbone, which achieve the best convergence results, respectively.
For training \diffucomet{} based on BART-large, we also follow Difformer \citep{gao2022difformer} to amplify the standard deviation of diffusion noise by a factor of $A=4$, \ie{}, to change the forward process as:
\begin{align}
q(\mathbf{z}_{t}|\mathbf{z}_{t-1})=\mathcal{N}(\mathbf{z}_{t};\sqrt{1-\beta_{t}}\mathbf{z}_{t-1},\beta_{t}A^{2}\mathbf{I})
\end{align}
where $t=1,2,...T$, which effectively avoids model collapse in training.
The total diffusion steps $T$ is set to $2000$.
We use AdamW \citep{loshchilov2018decoupled} as our training optimizer, with learning rate $1e^{-5}$ and no weight decay.
A linear learning rate scheduler is adopted with warm-up steps $2000$ and total training steps $150000$ and $200000$ for models based on BART-base (139M) and BART-large (406M), respectively.
We train our base-scale \diffucomet{} on 4 Tesla V100-SXM2 (32GB) GPUs with batch size set to $4$, while for large-scale \diffucomet{}, we use 4 NVIDIA A100-SXM4 (40GB) GPUs, with batch size set to $2$ instead.
15 and 36 hours are required to train base-scale and large-scale \diffucomet{} models, respectively.

For the pre-training of our fact embedding module with loss described in Eq.~(\ref{eq:embed_loss}), we adopt the same hyperparameter setting as training our diffusion module, except for learning rate changed to $2e^{-6}$ and batch size set to $128$ and $64$ for base-scale and large-scale models, respectively.
For pre-training large-scale (\ie{}, BART-large) fact embedding module, we add a weight decay of $0.01$, which leads to better convergence.
In \diffucomet{}-Entity, the two diffusion modules trained for generating contextual relevant head and tail entities share the same pre-trained entity embedding module.

\section{Evaluation Metrics}
\subsection{Clustering and Similarity Function}
\label{appendix:similarity_alignment}
For our evaluation based on fact clustering \wrt{} edit distance, we define the similarity function in our Alignment metric as $sim(\hat{k}_{i},k_{j})=1-Edit(\hat{k}_{i},k_{j})/MaxLen(\hat{k}_{i},k_{j})$, where $Edit$ denotes the word-level edit distance of two facts, and $MaxLen$ denotes the length of the longer fact of the two, \ie{}, the maximum possible edit distance for normalization.
Our distance measure for clustering also adopts the normalized edit distance, \ie{}, $Edit/MaxLen$.
For evaluation based on fact clustering \wrt{} Sentence-BERT embedding, we define the similarity function in our Alignment metric as $sim(\hat{k}_{i},k_{j})=max(CoS(\hat{k}_{i},k_{j}),0)$, where $CoS$ denotes the cosine similarity of two facts' Sentence-BERT embeddings.
We assume that facts with opposite meanings, \ie{}, negative similarity, are not considered as aligned with each other, so we cut off the negative values of cosine similarity.
While for the distance measure of clustering, we use the Euclidean distance of two facts' embeddings instead, which is typically adopted in DBSCAN \citep{ester1996density} clustering algorithm.

\begin{figure}[t]
\centering
\includegraphics[width=1.0\columnwidth]{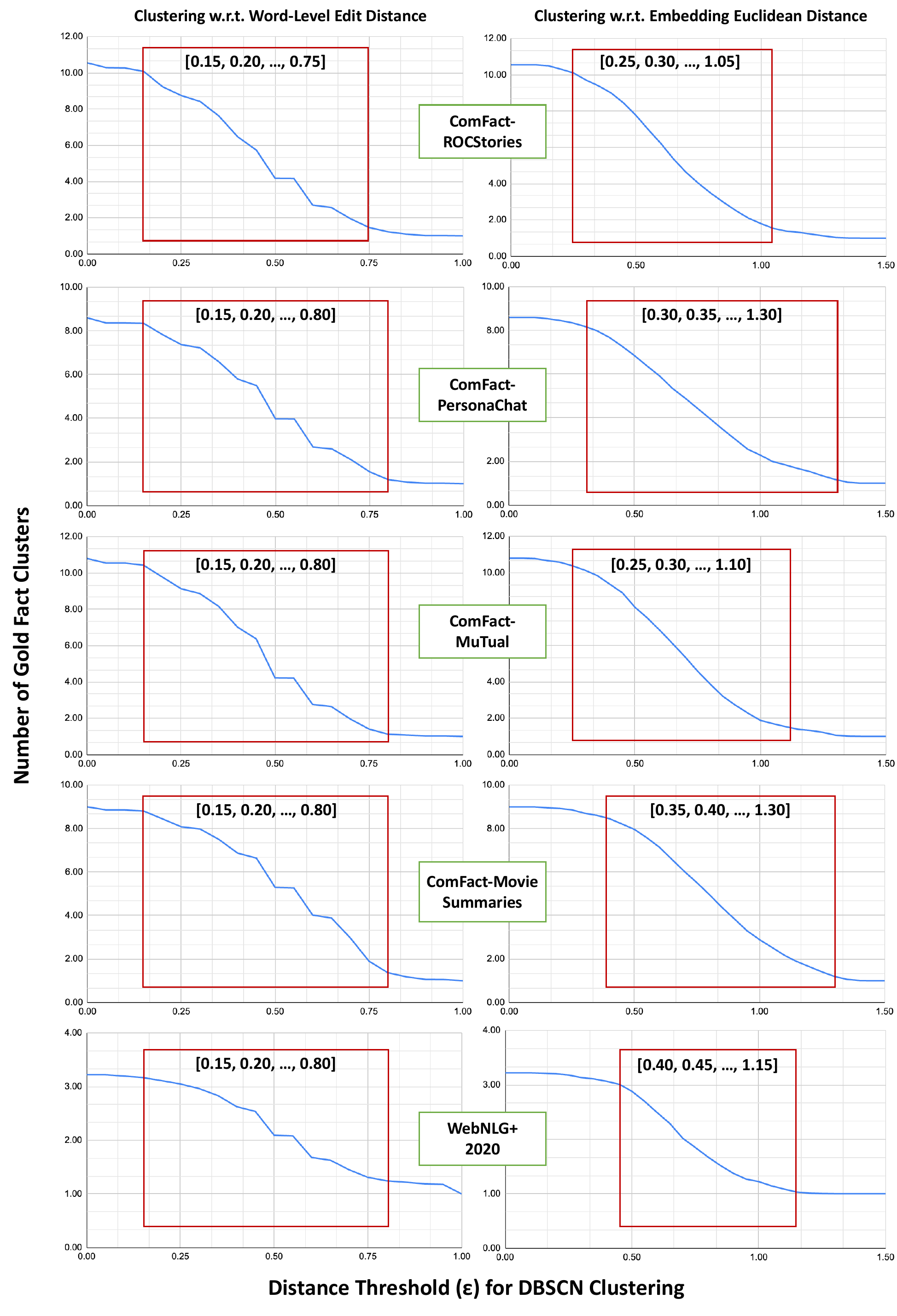}
\caption{Range selection (red square) of DBSCN clustering thresholds for our proposed metrics.}
\label{clustering_threshold}
\end{figure}

\subsection{Clustering Threshold Selection}
\label{appendix:clustering_threshold}
For our proposed clustering-based metrics as described in Section~\ref{sec:cluster_metrics}, we use DBSCAN \citep{ester1996density} algorithm to group facts into clusters.
To avoid bias on a specific clustering granularity, we consider a range of DBSCAN thresholds and take the average evaluation results across all thresholds in the range.
We consider a range with equal interval of $0.05$, where the number of gold fact clusters significantly changes from near the maximum (\ie{}, each fact as a cluster) to near the minimum (\ie{}, all facts grouped into one cluster).
Figure~\ref{clustering_threshold} shows the number of gold clusters as a function of the DBSCAN clustering threshold, and our selection of threshold ranges (red square) on each dataset.

\subsection{WebNLG Metrics}
\label{appendix:webnlg_metrics}
In the evaluation of WebNLG 2020 Challenge \citep{ferreira20202020}, each generated RDF fact (\ie{}, subject-predicate-object triple) is paired to a gold reference to compute its precision, recall and F1 based on named entity matching \citep{segura2013semeval}.
Three types of matching criterias are considered, including: a) each named entity in generated RDF needs to exactly match an entity in gold reference in order to be counted as true-positive, while its type in the RDF (\ie{}, whether it is in subject, predicate or object) does not need to match (\textbf{Exact Match}), b) each entity in generated RDF only needs to partially match an entity in gold reference, and its type does not matter (\textbf{Partial Match}), and c) each named entity in generated RDF needs to exactly match an entity in gold reference, and its type also needs to match (\textbf{Strict Match}).
For each matching criteria, optimal pairing (with the highest F1 score) between generated facts and gold references is searched by enumerating all possible permutations.
We report Strict Match scores in the main body of our paper in Table~\ref{tab:main_results_webnlg}, and include all three kinds of match scores in Table~\ref{tab:results_official_full_webnlg}.

\section{Data Preprocessing}
\label{appendix:data_preprocess}
\comfact{} \citep{gao2022comfact} benchmark contains social commonsense knowledge linked from \atomicTT{} \citep{hwang2021comet} knowledge base, which contains $\sim$1.33M facts covering physical entities, daily events and social interactions.
\atomicTT{} commonsense relations considered in our experiments are listed in Table~\ref{tab:relation_atomic}.
We preprocess \comfact{} and \atomicTT{} to filter out facts that have invalid tail entity ``none'' or contain fillable blank ``\_\_\_'', \ie{}, we do not consider facts with relation ``IsFilledBy''.
After preprocessing, $\sim$972K facts are involved in the training of our fact embedding and diffusion modules.
The original \comfact{} training data in the ROCStories portion only has $\sim1K$ contexts with gold annotations of relevant facts.
Due to the limited supervised data, we augment the training data with $\sim50K$ additional contexts sampled from the ROCStories corpus, and use a DeBERTa \citep{he2020deberta} fact linker developed from the \comfact{} benchmark to extract silver annotations of relevant facts from \atomicTT{} to each additional context.

For preprocessing WebNLG+ 2020 \citep{ferreira20202020} dataset, we follow Grapher \citep{melnyk2022knowledge} to remove underscores and surrounding quotes appeared in the dataset, and convert non-English characters into their closest available English characters, \eg{}, ``\~{o}'' and ``\r{a}'' are mapped to ``o'' and ``a''.
After preprocessing, We develop our models based on the $\sim35K$ WebNLG training texts and their linked RDF facts.

\begin{table}[t]
\centering
\resizebox{1.0\columnwidth}{!}{
\smallskip\begin{tabular}{lll}
\toprule
\textbf{Type} & \textbf{Relation} & \textbf{Relation Description} \\
\toprule
\multirow{6}*{Physical-} & ObjectUse & used for \\
\multirow{6}*{Entity} & AtLocation  & located or found at/in/on \\
        & MadeUpOf    & made (up) of \\
        & HasProperty & can be characterized by being/having \\
        & CapableOf   & is/are capable of \\
        & Desires     & desires \\
        & NotDesires  & do(es) not desire \\
\midrule
\multirow{6}*{Event} & IsAfter     & happens after \\
        & IsBefore    & happens before \\
        & HasSubEvent & includes the event/action \\
        & HinderedBy  & can be hindered by \\
        & Causes      & causes \\
        & xReason     & because \\
\midrule
\multirow{8}*{Social-}     & xNeed       & but before, person X needs \\
\multirow{8}*{Interaction} & xAttr       & person X is seen as \\
        & xEffect     & as a result, person X will \\
        & xReact      & as a result, person X feels \\
        & xWant       & as a result, person X wants \\
        & xIntent     & because person X wants \\  
        & oEffect     & as a result, others will \\
        & oReact      & as a result, others feel \\
        & oWant       & as a result, others want \\
\bottomrule
\end{tabular}
}
\caption{Commonsense relations in \atomicTT{} knowledge base that are considered in our experiments on \comfact{} benchmark.}
\label{tab:relation_atomic}
\end{table}

\begin{table*}[t]
\centering
\resizebox{1.0\textwidth}{!}{
\smallskip\begin{tabular}{clccccccccc}
\hline
\multirow{2}*{\textbf{Backbone}} & \multirow{2}*{\textbf{Model}} & \multirow{2}*{\textbf{\# Facts}}  & \multicolumn{4}{c}{\textbf{Clustering \wrt{} Word-Level Edit Distance}} & \multicolumn{4}{c}{\textbf{Clustering \wrt{} Embedding Euclidean Distance}}\\
            \cmidrule(lr){4-7} \cmidrule(lr){8-11}
 &  &  & \textbf{\# Clusters} & \textbf{Relevance} & \textbf{Alignment} & \textbf{RA-F1} & \textbf{\# Clusters} & \textbf{Relevance} & \textbf{Alignment} & \textbf{RA-F1} \\
\hline
\multirow{7}*{BART} & Greedy        & 2.48  & 1.08 & 32.04 & 31.98 & 32.01 & 1.09 & 32.11 & 48.59 & 38.67 \\
\multirow{7}*{(base)} & Sampling-10   & 10.00 & 5.59 & 39.20 & 46.03 & 42.34 & 5.64 & 38.93 & 64.51 & 48.56 \\
        & Sampling-15   & 15.00 & 7.64 & 37.18 & 49.78 & 42.57 & 7.82 & 36.86 & 68.00 & 47.81 \\
        & Beam-10       & 10.00 & 2.63 & 38.30 & 44.96 & 41.36 & 2.83 & 38.87 & 59.58 & 47.05 \\
        & Beam-15       & 15.00 & 3.48 & 41.46 & 48.04 & 44.51 & 3.97 & 42.88 & 63.14 & 51.07 \\
        \cmidrule(lr){2-11}
        & \diffucomet-Fact   & 13.40 & 4.74 & 59.75 & 54.07 & 56.77 & 5.85 & 60.32 & 73.38 & 66.21 \\
        % & ~~~- Joint-Train & 3.64  & 2.05 & \textbf{73.01} & 42.54 & 53.76 & 2.09 & \textbf{73.19} & 56.84 & 63.99 \\
        & \diffucomet-Entity & 10.08 & 4.51 & 62.27 & 54.61 & 58.19 & 5.24 & 61.77 & 71.54 & 66.30 \\
\midrule
\multirow{6}*{BART} & Greedy & 2.20  & 1.38 & 60.45 & 36.11 & 45.21 & 1.37 & 60.22 & 52.31 & 55.99 \\
\multirow{6}*{(large)} & Sampling-10 & 10.00 & 6.68 & 56.09 & 52.10 & 54.02 & 6.40 & 56.68 & 73.86 & 64.14 \\
    & Sampling-15   & 15.00 & \textbf{8.89} & 56.24 & 55.18 & 55.70 & \textbf{8.56} & 56.57 & 76.30 & 64.97 \\
    & Beam-10       & 10.00 & 3.32 & 64.94 & 50.72 & 56.96 & 3.51 & 64.37 & 69.14 & 66.67 \\
    & Beam-15       & 15.00 & 4.17 & 64.18 & 53.66 & 58.45 & 4.60 & 64.35 & 71.35 & 67.67 \\
    \cmidrule(lr){2-11}
    & \diffucomet-Fact   & 12.88 & 4.47 & 65.82 & 54.18 & 59.44 & 5.24 & 65.64 & 71.65 & 68.51 \\
    & \diffucomet-Entity & 12.89 & 5.09 & 67.00 & 58.22 & \textbf{62.30} & 5.67 & 66.39 & 74.38 & \textbf{70.16}  \\
\midrule
\multirow{4}*{\comet{}-} & Greedy & 1.96 & 1.14 & 61.27 & 34.76 & 44.36 & 1.19 & 61.42 & 50.64 & 55.51 \\
\multirow{4}*{BART} & Sampling-10 & 10.00 & 6.45 & 56.79 & 53.36 & 55.02 & 6.30 & 56.60 & 73.64 & 64.01 \\
    & Sampling-15   & 15.00 & 8.52 & 55.78 & \textbf{58.99} & 57.34 & 8.39 & 56.19 & \textbf{77.97} & 65.31 \\
    & Beam-10       & 10.00 & 3.78 & 65.62 & 53.45 & 58.91 & 3.89 & 65.73 & 70.65 & 68.10 \\
    & Beam-15       & 15.00 & 4.78 & 64.91 & 54.77 & 59.41 & 5.09 & 65.03 & 71.64 & 68.18 \\
\midrule
T5 (large) & Grapher & 5.08  & 1.75 & 67.82 & 33.07 & 44.46 & 2.60 & 68.29 & 40.58 & 50.91 \\
\midrule
- & Gold          & 10.55 & 5.64 & 81.06 & - & - & 5.64 & 80.90 & - & - \\
\hline
\end{tabular}
}
\caption{Clustering-based evaluation results on the \textbf{ROCStories} portion of \comfact{}. Best results (excluding Gold references) are in bold. Different numbers after Sampling and Beam denote various sampling numbers or beam search sizes being tested.}
\label{tab:results_cluster_full_roc}
\end{table*}

\begin{table}[t]
\centering
\resizebox{1.0\columnwidth}{!}{
\smallskip\begin{tabular}{@{~}clc@{~~~}c@{~~~}c@{~~~}c@{~}}
\hline
\textbf{Backbone} & \textbf{Model} & \textbf{Distinct-4}  & \textbf{BLEU} & \textbf{METEOR} & \textbf{ROUGE-L} \\
\hline
\multirow{6}*{BART} & Greedy        & \textbf{99.90} & 8.70 & 40.49 & 44.43 \\
\multirow{6}*{(base)} & Sampling-10   & 85.29 & 7.16 & 37.78 & 39.20 \\
       & Sampling-15   & 81.57 & 8.24 & 38.35 & 40.13 \\
       & Beam-10       & 50.32 & 12.25 & 42.23 & 43.53 \\
       & Beam-15       & 45.21 & 11.51 & 42.04 & 42.91 \\
       \cmidrule(lr){2-6}
      & \diffucomet-Fact   & 57.87 & 12.09 & 46.43 & 47.13 \\
      & \diffucomet-Entity & 70.02 & 14.25 & 43.34 & 45.08 \\
\midrule
\multirow{6}*{BART} & Greedy & 93.01 & 9.12 & 43.98 & 46.26 \\
\multirow{6}*{(large)} & Sampling-10 & 86.33 & 9.89 & 43.85 & 43.69 \\
    & Sampling-15   & 81.56 & 9.47 & 43.28 & 43.15 \\
    & Beam-10       & 47.03 & 15.02 & 48.56 & 48.15 \\
    & Beam-15       & 43.73 & 13.11 & 47.70 & 46.35 \\
    \cmidrule(lr){2-6}
    & \diffucomet-Fact   & 52.46 & 15.98 & 50.06 & 51.44 \\
    & \diffucomet-Entity & 63.49 & 17.01 & 47.61 & 48.40 \\
\midrule
\multirow{4}*{\comet{}-} & Greedy & 65.95 & 18.01 & \textbf{52.32} & \textbf{54.96} \\
\multirow{4}*{BART} & Sampling-10 & 83.29 & 13.35 & 44.77 & 45.80 \\
    & Sampling-15   & 79.01 & 12.69 & 44.43 & 45.58 \\
    & Beam-10       & 51.13 & \textbf{19.89} & 50.14 & 50.48 \\
    & Beam-15       & 47.27 & 16.97 & 47.39 & 47.19 \\
\midrule
T5 (large) & Grapher & 67.83  & 1.40 & 23.96 & 27.21 \\
\midrule
  -  &      Gold     & 80.45 & - & - & - \\
\hline
\end{tabular}
}
\caption{Evaluation results of natural language generation metrics on the \textbf{ROCStories} portion of \comfact{}. Notations are same as Table~\ref{tab:results_cluster_full_roc}.}
\label{tab:results_nlg_full_roc}
\end{table}

\begin{figure*}
\centering
\includegraphics[width=\textwidth]{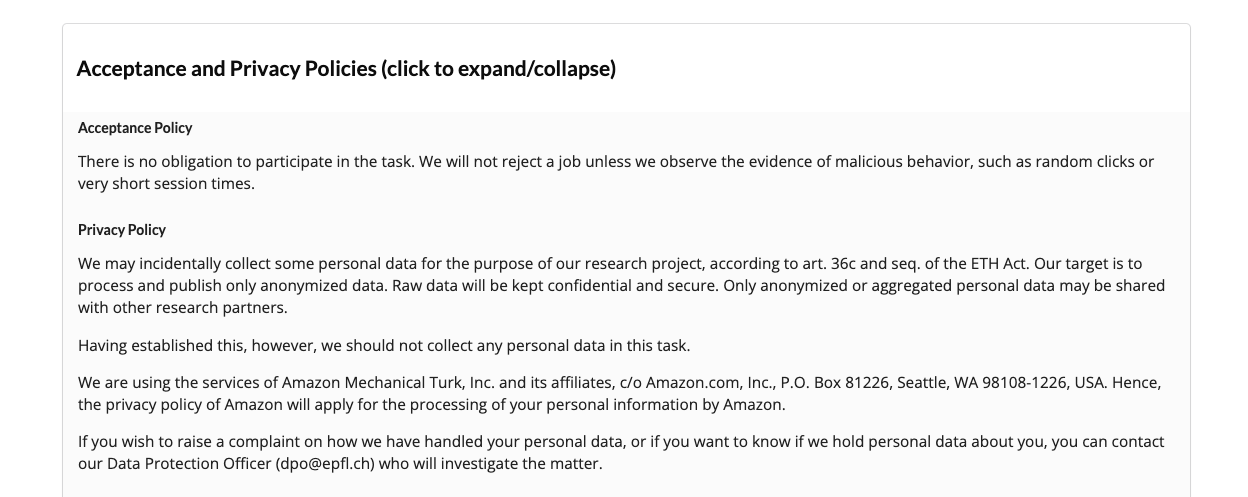}
\caption{Screenshot of Amazon MTurk Acceptance and Privacy Policy}
\label{fig:mturk-policy}
\end{figure*}

\begin{figure*}
\centering
\includegraphics[width=\textwidth]{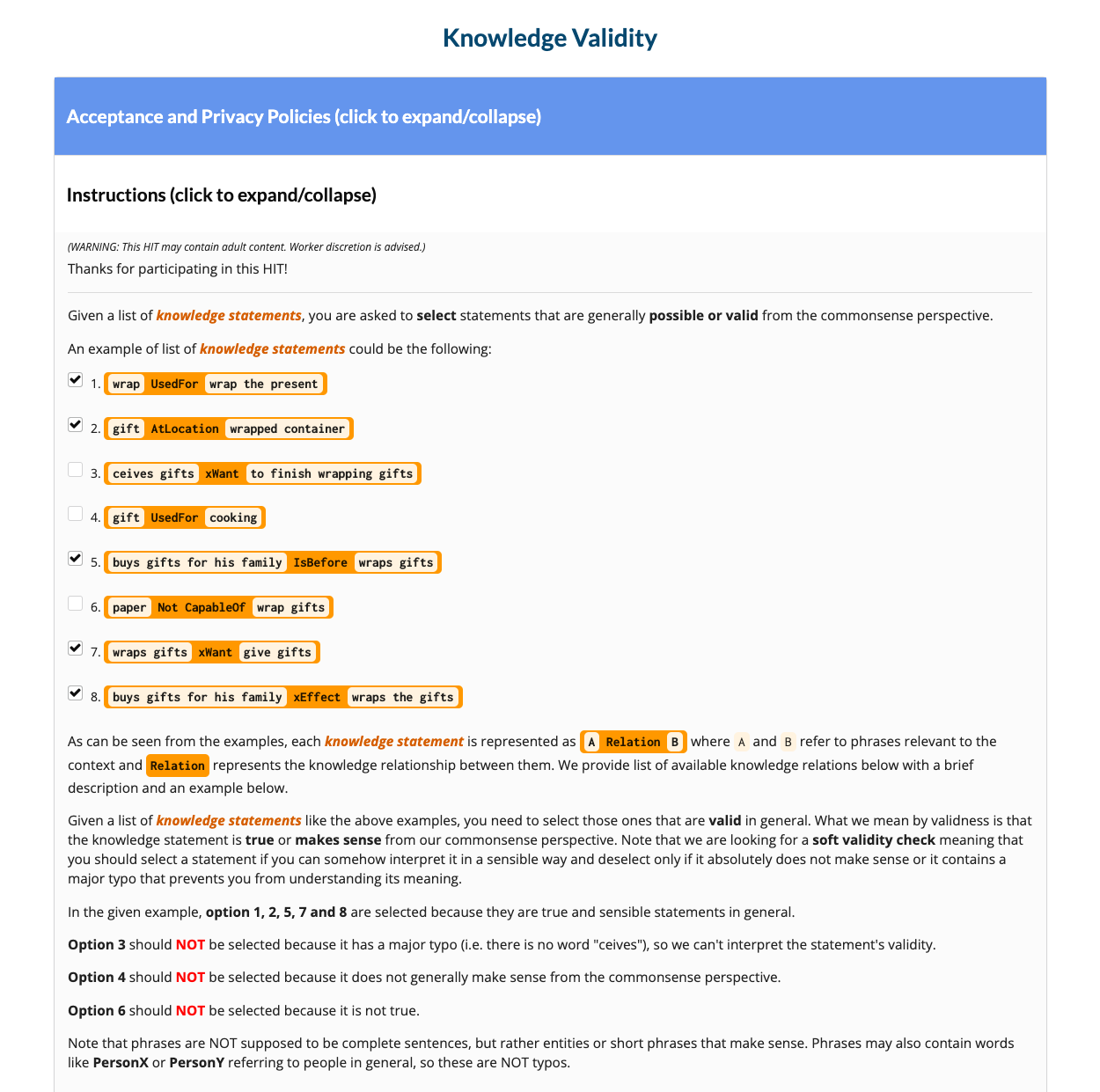}
\caption{Screenshot of Amazon MTurk instructions for knowledge validation task.}
\label{fig:mturk-validity}
\end{figure*}

\begin{figure*}
\centering
\includegraphics[width=\textwidth]{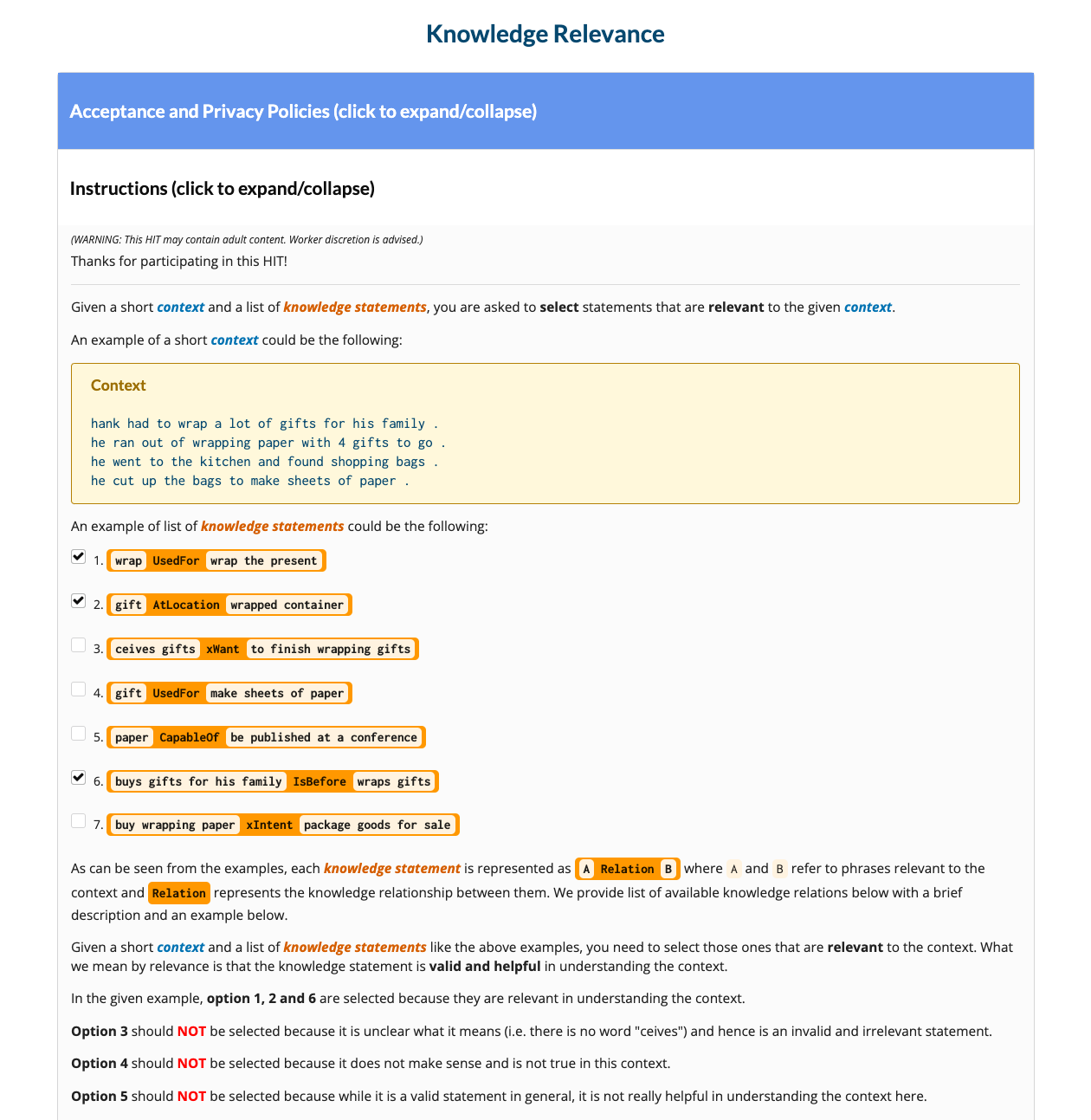}
\caption{Screenshot of Amazon MTurk instructions for knowledge relevance task.}
\label{fig:mturk-relevance}
\end{figure*}

% \begin{table}[t]
% \centering
% \resizebox{1.0\columnwidth}{!}{
% \smallskip\begin{tabular}{lcc}
% \hline
% \textbf{Model} & \textbf{Validity} & \textbf{Relevance} \\
% \hline
% Sampling-\comet{} & 49.45 & 30.20 \\
% Beam-\comet{} & \textbf{74.80} & 42.81 \\
% % Grapher & 34.86 & 32.27 \\
% \midrule
% \diffucomet-Fact & 70.00 & \underline{48.27} \\
% \diffucomet-Entity & \underline{74.15} & \textbf{54.18} \\
% \midrule
% Gold & 94.79 & 82.04 \\
% \hline
% \end{tabular}
% }
% \caption{Human evaluation results on the ROCStories portion of \comfact{}. Notations are same as Table~\ref{tab:main_results_roc}.}
% \label{tab:human_eval}
% \end{table}

\begin{table*}[t]
\centering
\resizebox{1.0\textwidth}{!}{
\smallskip\begin{tabular}{clccccccccccc}
\hline
\multirow{2}*{\textbf{Backbone}} & \multirow{2}*{\textbf{Model}} & \multirow{2}*{\textbf{\# Facts}}  & \multicolumn{4}{c}{\textbf{Clustering \wrt{} Word-Level Edit Distance}} & \multicolumn{4}{c}{\textbf{Clustering \wrt{} Embedding Euclidean Distance}}\\
            \cmidrule(lr){4-7} \cmidrule(lr){8-11}
 &  &  & \textbf{\# Clusters} & \textbf{Relevance} & \textbf{Alignment} & \textbf{RA-F1} & \textbf{\# Clusters} & \textbf{Relevance} & \textbf{Alignment} & \textbf{RA-F1} \\
\hline
\multirow{4}*{BART} & Greedy        & 2.62 & 1.24 & 33.52 & 35.64 & 34.55 & 1.24 & 33.57 & 48.61 & 39.71 \\
\multirow{4}*{(base)} & Sampling-10   & 10.00 & \textbf{5.23} & 29.35 & 44.49 & 35.37 & 4.63 & 28.26 & 58.85 & 38.18 \\
        & Beam-15       & 15.00 & 3.65 & 31.11 & 47.01 & 37.44 & 3.49 & 30.41 & 58.90 & 40.11 \\
        & \diffucomet-Fact   & 13.73 & 4.97 & 44.25 & 52.81 & 48.15 & 5.24 & 44.76 & \textbf{69.24} & 54.37 \\
        & \diffucomet-Entity & 11.40 & 4.99 & 50.39 & 54.84 & 52.52 & \textbf{4.94} & 49.36 & 68.55 & 57.39 \\
\midrule
\multirow{2}*{BART} & Beam-15       & 15.00 & 4.44 & 53.98 & 54.13 & 54.05 & 4.04 & 54.07 & 63.17 & 58.27 \\
\multirow{2}*{(large)} & \diffucomet-Fact   & 10.82 & 4.48 & \textbf{55.44} & 55.20 & 55.32 & 3.89 & \textbf{55.02} & 65.20 & 59.68 \\
    & \diffucomet-Entity & 12.06 & 4.72 & 55.08 & \textbf{57.12} & \textbf{56.08} & 4.48 & 54.42 & 68.11 & \textbf{60.50}  \\
\midrule
\comet{}-BART & Beam-15    & 15.00 & 4.86 & 54.02 & 54.78 & 54.40 & 4.27 & 53.97 & 65.15 & 59.04 \\
T5 (large) & Grapher & 4.53 & 1.68 & 47.74 & 30.51 & 37.23 & 1.57 & 47.94 & 36.18 & 41.24 \\
\midrule
- & Gold          & 8.60 & 4.76 & 70.42 & - & - & 4.28 & 70.42 & - & - \\
\hline
\end{tabular}
}
\caption{Zero-shot clustering-based evaluation results on the \textbf{PersonaChat} portion of \comfact{}. Notations are same as Table~\ref{tab:results_cluster_full_roc}.}
\label{tab:results_cluster_full_persona}
\end{table*}

\begin{table}[t]
\centering
\resizebox{1.0\columnwidth}{!}{
\smallskip\begin{tabular}{@{~}clc@{~~~}c@{~~~}c@{~~~}c@{~}}
\hline
\textbf{Backbone} & \textbf{Model} & \textbf{Distinct-4}  & \textbf{BLEU} & \textbf{METEOR} & \textbf{ROUGE-L} \\
\hline
\multirow{4}*{BART} & Greedy        & \textbf{97.83} & 8.72 & 44.44 & 46.52 \\
\multirow{4}*{(base)} & Sampling-10   & 86.81 & 4.09 & 32.95 & 33.80 \\
        & Beam-15       & 53.05 & 8.06 & 37.43 & 38.62 \\
        & \diffucomet-Fact   & 63.40 & 5.84 & 37.53 & 39.33 \\
        & \diffucomet-Entity & 73.38 & 9.04 & 34.35 & 36.46 \\
\midrule
\multirow{2}*{BART}     & Beam-15       & 47.64 & 8.71 & 41.40 & 40.44 \\
\multirow{2}*{(large)} & \diffucomet-Fact   & 57.23 & 8.05 & \textbf{45.83} & \textbf{47.11} \\
       & \diffucomet-Entity & 68.54 & \textbf{11.11} & 38.88 & 40.04 \\
\midrule
\comet{}-BART & Beam-15       & 50.13 & 10.25 & 43.47 & 42.38 \\
T5 (large) & Grapher & 52.99 & 0.68 & 19.91 & 22.41 \\
\midrule
  -  &      Gold     & 84.96 & - & - & - \\
\hline
\end{tabular}
}
\caption{Zero-shot evaluation results of natural language generation metrics on the \textbf{PersonaChat} portion of \comfact{}. Notations are same as Table~\ref{tab:results_cluster_full_roc}.}
\label{tab:results_nlg_full_persona}
\end{table}

\begin{table*}[t]
\centering
\resizebox{1.0\textwidth}{!}{
\smallskip\begin{tabular}{clccccccccccc}
\hline
\multirow{2}*{\textbf{Backbone}} & \multirow{2}*{\textbf{Model}} & \multirow{2}*{\textbf{\# Facts}}  & \multicolumn{4}{c}{\textbf{Clustering \wrt{} Word-Level Edit Distance}} & \multicolumn{4}{c}{\textbf{Clustering \wrt{} Embedding Euclidean Distance}}\\
            \cmidrule(lr){4-7} \cmidrule(lr){8-11}
 &  &  & \textbf{\# Clusters} & \textbf{Relevance} & \textbf{Alignment} & \textbf{RA-F1} & \textbf{\# Clusters} & \textbf{Relevance} & \textbf{Alignment} & \textbf{RA-F1} \\
\hline
\multirow{4}*{BART} & Greedy    & 2.50 & 1.16 & 35.85 & 33.63 & 34.70 & 1.19 & 36.10 & 48.53 & 41.40 \\
\multirow{4}*{(base)} & Sampling-10   & 10.00 & \textbf{5.68} & 43.77 & 45.76 & 44.74 & \textbf{5.88} & 43.98 & 63.75 & 52.05 \\
        & Beam-15       & 15.00 & 3.55 & 41.27 & 49.72 & 45.10 & 4.08 & 42.27 & 63.17 & 50.65 \\
        & \diffucomet-Fact   & 13.11 & 4.54 & 57.64 & 52.25 & 54.81 & 5.57 & 57.51 & 70.10 & 63.18 \\
        & \diffucomet-Entity & 10.63 & 4.60 & 60.08 & 54.65 & 57.24 & 5.27 & 59.08 & 68.88 & 63.60 \\
\midrule
\multirow{2}*{BART} & Beam-15    & 15.00 & 3.92 & 64.19 & 51.75 & 57.30 & 4.45 & 62.41 & 67.31 & 64.77 \\
\multirow{2}*{(large)} & \diffucomet-Fact  & 10.46 & 4.33 & 64.74 & 54.51 & 59.19 & 4.80 & 64.13 & 68.07 & 66.04 \\
    & \diffucomet-Entity & 11.85 & 4.70 & 64.39 & \textbf{55.91} & \textbf{59.85} & 5.39 & 63.82 & \textbf{71.22} & \textbf{67.32} \\
\midrule
\comet{}-BART & Beam-15   & 15.00 & 4.52 & 61.88 & 54.04 & 57.69 & 4.75 & 60.56 & 69.72 & 64.82 \\
T5 (large) & Grapher & 4.50 & 1.70 & \textbf{73.30} & 32.74 & 45.26 & 1.78 & \textbf{73.33} & 43.13 & 54.31 \\
\midrule
- & Gold          & 10.80 & 5.58 & 74.63 & - & - & 5.79 & 74.77 & - & - \\
\hline
\end{tabular}
}
\caption{Zero-shot clustering-based evaluation results on the \textbf{MuTual} portion of \comfact{}. Notations are same as Table~\ref{tab:results_cluster_full_roc}.}
\label{tab:results_cluster_full_mutual}
\end{table*}

\begin{table}[t]
\centering
\resizebox{1.0\columnwidth}{!}{
\smallskip\begin{tabular}{@{~}clc@{~~~}c@{~~~}c@{~~~}c@{~}}
\hline
\textbf{Backbone} & \textbf{Model} & \textbf{Distinct-4}  & \textbf{BLEU} & \textbf{METEOR} & \textbf{ROUGE-L} \\
\hline
\multirow{4}*{BART} & Greedy         & \textbf{97.47} & \textbf{14.05} & \textbf{49.89} & 50.78 \\
\multirow{4}*{(base)} & Sampling-10  & 86.42 & 5.61 & 37.11 & 38.60 \\
        & Beam-15       & 49.31 & 11.57 & 45.47 & 45.51 \\
        & \diffucomet-Fact   & 60.66 & 8.71 & 44.23 & 46.11 \\
        & \diffucomet-Entity & 70.94 & 11.08 & 40.28 & 42.15 \\
\midrule
\multirow{2}*{BART}     & Beam-15       & 43.91 & 11.37 & 46.86 & 46.75 \\
\multirow{2}*{(large)} & \diffucomet-Fact   & 52.00 & 12.33 & 49.50 & \textbf{50.97} \\
       & \diffucomet-Entity & 66.12 & 12.68 & 45.11 & 45.73 \\
\midrule
\comet{}-BART & Beam-15       & 47.40 & 12.40 & 49.12 & 48.57 \\
T5 (large) & Grapher & 51.30 & 1.96 & 24.70 & 29.36\\
\midrule
  -  &      Gold     & 80.99 & - & - & - \\
\hline
\end{tabular}
}
\caption{Zero-shot evaluation results of natural language generation metrics on the \textbf{MuTual} portion of \comfact{}. Notations are same as Table~\ref{tab:results_cluster_full_roc}.}
\label{tab:results_nlg_full_mutual}
\end{table}

\begin{table*}[t]
\centering
\resizebox{1.0\textwidth}{!}{
\smallskip\begin{tabular}{clccccccccccc}
\hline
\multirow{2}*{\textbf{Backbone}} & \multirow{2}*{\textbf{Model}} & \multirow{2}*{\textbf{\# Facts}}  & \multicolumn{4}{c}{\textbf{Clustering \wrt{} Word-Level Edit Distance}} & \multicolumn{4}{c}{\textbf{Clustering \wrt{} Embedding Euclidean Distance}}\\
            \cmidrule(lr){4-7} \cmidrule(lr){8-11}
 &  &  & \textbf{\# Clusters} & \textbf{Relevance} & \textbf{Alignment} & \textbf{RA-F1} & \textbf{\# Clusters} & \textbf{Relevance} & \textbf{Alignment} & \textbf{RA-F1} \\
\hline
\multirow{4}*{BART} & Greedy  & 2.59 & 1.12 & 33.50 & 25.28 & 28.82 & 1.11 & 33.38 & 37.95 & 35.52 \\
\multirow{4}*{(base)} & Sampling-10   & 10.00 & 4.90 & 26.61 & 33.31 & 29.59 & 4.24 & 24.96 & 50.26 & 33.36 \\
        & Beam-15      & 15.00 & 3.54 & 30.45 & 36.07 & 33.02 & 3.17 & 29.13 & 49.63 & 36.71 \\
        & \diffucomet-Fact   & 14.61 & 6.02 & 35.97 & 38.76 & 37.31 & 5.49 & 36.29 & 59.83 & 45.18 \\
        & \diffucomet-Entity & 15.82 & \textbf{6.31} & 39.86 & 39.93 & 39.89 & 5.76 & 39.57 & 57.55 & 46.90 \\
\midrule
\multirow{2}*{BART} & Beam-15    & 15.00 & 4.46 & 42.92 & 32.79 & 37.18 & 3.70 & 42.52 & 50.46 & 46.15 \\
\multirow{2}*{(large)} & \diffucomet-Fact  & 8.29 & 3.01 & 41.50 & 30.47 & 35.14 & 2.89 & 40.82 & 46.59 & 43.51 \\
    & \diffucomet-Entity & 13.50 & 6.28 & 44.08 & \textbf{40.46} & \textbf{42.19} & \textbf{5.84} & 42.70 & \textbf{61.56} & \textbf{50.42} \\
\midrule
\comet{}-BART & Beam-15   & 15.00 & 5.06 & 41.97 & 34.63 & 37.95 & 4.04 & 41.54 & 51.24 & 45.88 \\
T5 (large) & Grapher & 5.34 & 1.83 & \textbf{54.09} & 23.74 & 33.00 & 1.50 & \textbf{54.12} & 34.27 & 41.97 \\
\midrule
- & Gold          & 9.00 & 5.64 & 58.55 & - & - & 4.81 & 58.37 & - & - \\
\hline
\end{tabular}
}
\caption{Zero-shot clustering-based evaluation results on the \textbf{MovieSummaries} portion of \comfact{}. Notations are same as Table~\ref{tab:results_cluster_full_roc}.}
\label{tab:results_cluster_full_movie}
\end{table*}

\begin{table}[t]
\centering
\resizebox{1.0\columnwidth}{!}{
\smallskip\begin{tabular}{@{~}clc@{~~~}c@{~~~}c@{~~~}c@{~}}
\hline
\textbf{Backbone} & \textbf{Model} & \textbf{Distinct-4}  & \textbf{BLEU} & \textbf{METEOR} & \textbf{ROUGE-L} \\
\hline
\multirow{4}*{BART} & Greedy         & \textbf{95.18} & 5.14 & 34.24 & 36.53 \\
\multirow{4}*{(base)} & Sampling-10  & 90.99 & 2.52 & 24.56 & 28.08 \\
        & Beam-15       & 53.65 & 4.82 & 28.33 & 31.56 \\
        & \diffucomet-Fact   & 63.93 & 2.27 & 27.09 & 30.08 \\
        & \diffucomet-Entity & 67.24 & 2.68 & 24.14 & 26.95 \\
\midrule
\multirow{2}*{BART}     & Beam-15       & 47.57 & 4.89 & 31.41 & 33.19 \\
\multirow{2}*{(large)} & \diffucomet-Fact   & 43.68 & \textbf{5.26} & \textbf{34.55} & \textbf{38.80} \\
       & \diffucomet-Entity & 67.13 & 3.83 & 26.63 & 29.32 \\
\midrule
\comet{}-BART & Beam-15       & 50.29 & 5.18 & 31.36 & 33.39 \\
T5 (large) & Grapher & 42.76 & 0.46 & 18.24 & 21.28 \\
\midrule
  -  &      Gold     & 87.39 & - & - & - \\
\hline
\end{tabular}
}
\caption{Zero-shot evaluation results of natural language generation metrics on the \textbf{MovieSummaries} portion of \comfact{}. Notations are same as Table~\ref{tab:results_cluster_full_roc}.}
\label{tab:results_nlg_full_movie}
\end{table}

\section{Human Evaluation Details}
\label{appendix:human_evaluation}
% In our human evaluation that checks the contextual relevance of generated facts, we also ask workers to judge the validity of each fact, \ie{}, whether a fact is always true or at least makes sense in certain circumstances.
% Same as the evaluation of relevance, we label a fact as valid if it passes the verification of two or all three workers.
% Invalid facts are also judged as irrelevant to the context.

Our annotator pool for human evaluation contains $58$ Amazon Mechanical Turk workers who are located in the USA and have been previously qualified by us for other similar tasks. To prepare the workers for the new tasks of assessing the validity and relevance of knowledge in a given context, we share the instructions with them beforehand and do a small pilot run where we evaluate the quality of the worker annotations and give feedback if needed. We pay each worker $\$0.10$ for each task. % which takes on average $40$ seconds to complete. 
Figure \ref{fig:mturk-policy}, \ref{fig:mturk-validity} and \ref{fig:mturk-relevance} show screenshots of our acceptance/privacy policy and instructions for knowledge validation and relevance tasks.
Our data collection protocol follows Amazon Mechanical Turk regulations, and is approved by our organization in terms of ethics.

\section{Full Results of Knowledge Generation}
\label{appendix:full_results}

\subsection{ROCStories}
\label{appendix:roc_stories}
In Table~\ref{tab:results_cluster_full_roc} and \ref{tab:results_nlg_full_roc}, we present our full evaluation results of contextual commonsense knowledge generation on the ROCStories portion of \comfact{} benchmark.
For evaluating Sampling and Beam baseline models, we test two sampling or beam search sizes around the average number of gold facts per context, \ie{}, $10$ and $15$ as indicated by the suffix numbers, and adopt the size which achieves better F1 results.
On both base and large model scales, \diffucomet{} models achieve consistently better balance between the diversity (\ie{}, \#~Clusters) and accuracy (\ie{}, RA-F1) of knowledge generation, compared to baseline models that typically perform generation in the autoregressive manner.

% Table~\ref{tab:human_eval} presents our human evaluation results, where knowledge generated by our \diffucomet{} models achieves comparable validity to the strongest baseline Beam-\comet{}, while outperforms the baseline in terms of contextual relevance.
% This again demonstrates the superior knowledge generation accuracy of \diffucomet{}.

\subsection{Context Generalization}
\label{appendix:context_generalization}
In this section, we present zero-shot evaluation results of models (trained on the contexts of ROCStories) generalizing to the contexts of other three \comfact{} portions, including PersonaChat (Table~\ref{tab:results_cluster_full_persona} and \ref{tab:results_nlg_full_persona}), MuTual (Table~\ref{tab:results_cluster_full_mutual} and \ref{tab:results_nlg_full_mutual}) and MovieSummaries (Table~\ref{tab:results_cluster_full_movie} and \ref{tab:results_nlg_full_movie}).

We observe that both \diffucomet{} models generalize well to the contexts of PersonaChat and MuTual, whose generated knowledge possesses comparable diversity (\ie{}, \#~Clusters) and better accuracy (\ie{}, RA-F1) than the strongest baseline model Beam-\comet{}.
More interestingly, we find that \diffucomet{}-Entity achieves larger points of improvements over baselines on the more challenging MovieSummaries-style contexts, while \diffucomet{}-Fact struggles to outperform the strongest baseline Beam-\comet{}, showing that entity-level diffusion is more robust to the shift of narrative contexts, likely due to the more fine-grained multi-step learning of context-to-knowledge mapping.

\subsection{Case Study and Knowledge Types}
\label{appendix:case_study}
Table~\ref{tab:examples_full} showcases the knowledge generation results of \diffucomet{} models in a narrative context sampled from \comfact{} ROCStories, compared to the sampling and beam search baselines Sample-\comet{} and Beam-\comet{}.
Facts that are novel (\ie{}, beyond the coverage of gold references) and relevant to the context are labeled in bold.
We find that both \diffucomet{}-Fact and \diffucomet{}-Entity can generate facts that are rich in diversity, covering both physical entities (\eg{}, baseball cap) and social events (\eg{}, go on vacation).
Novel facts generated by \diffucomet{} models also uncover implicit inter-connections between entities or events in the narrative context, \eg{}, ``vacation'' and ``family'' are associated because ``X goes on vacation'' to ``spend time with family''.
By contrast, Beam-\comet{} model mainly generates simple facts about physical entities, and Sample-\comet{} model generates many facts that are irrelevant to the context, \eg{}, ``field is used for playing baseball''.

We also conduct a study on the proportion of different knowledge types that each model generates per context, based on the ROCStories portion of \comfact{} benchmark.
In particular, we divide commonsense facts into three types according to their relation groups under \atomicTT{} knowledge scheme, as shown in Table~\ref{tab:relation_atomic}, including facts that are centered on physical entities, events and social interactions.
Table~\ref{tab:knowledge_type} shows the results of knowledge proportion generated by \diffucomet{} and baseline models, with gold references.
Compared to Sampling-\comet{} and Beam-\comet{} baselines, \diffucomet{} models generate a larger proportion of facts that reveal complex event or social inter-connections.
The proportion of social-based facts generated by \diffucomet{} even significantly surpasses the gold references.
All above results imply that diffusion models have the potential to uncover more in-depth and implicit commonsense inferences from narrative contexts, which may not be easily extracted from existing knowledge bases.

\subsection{WebNLG+ 2020}
\label{appendix:webnlg_results}
We present our full evaluation results on the WebNLG+ 2020 benchmark in Table~\ref{tab:results_cluster_full_webnlg}, \ref{tab:results_nlg_full_webnlg} and \ref{tab:results_official_full_webnlg}.
For evaluating Sampling and Beam baselines, we set both sampling and beam search sizes as $5$, which is around the average number of gold facts per context.
Consistent with the evaluation results on \comfact{}, \diffucomet{} models keep achieving better performances on the WebNLG task of factual knowledge generation, implying that our method of diffusion-based contextual knowledge generation can generalize well to knowledge beyond commonsense.

\subsection{Comparison of Fact and Entity Diffusion}
\label{appendix:comparison_fact_entity}
For the comparison in between our two diffusion models, \diffucomet-Entity in general outperforms \diffucomet-Fact on our proposed metrics, which may benefit from more fine-grained multi-step learning of knowledge construction in pipeline.
However, \diffucomet-Fact is computational cheaper, \ie{}, only requires a single step of fact diffusion instead of two steps of (head and tail) entity diffusion and a relation prediction.
% \diffucomet-Entity requires more complex training and inference, \ie{}, two steps of (head and tail) entity generation and a relation prediction, while \diffucomet-Fact is computational cheaper.

\section{Claim of Usage}
Our use of existing scientific artifacts cited in this paper is consistent with their intended use.
Our developed code and models are intended to be used for only research purposes, any usage of our scientific artifacts that is outside of research contexts should not be allowed.

\begin{table}[t]
\centering
\resizebox{1.0\columnwidth}{!}{
\smallskip\begin{tabular}{cl}
\toprule
% \multicolumn{2}{c}{\multirow{5}*{Narrative Context}}
% \midrule
\multirow{4}*{Narrative}    & Dustin loved to wear his baseball cap everywhere he went. \\
\multirow{4}*{Context}      & On vacation his family visited the windy city of Chicago. \\
                            & Dustin's baseball cap blew off his head and into the street. \\
                            & His dad waited until it was safe before getting Dustin's cap. \\
                            & He loved his baseball cap even though it was a little dirty. \\
\midrule
\multirow{10}*{Gold}    & cap, used for, to wear on head \\
                        & cap, used for, wear on their heads \\
                        & head cap, used for, put on head \\
                        & vacation, used for, have fun on  \\
                        & vacation, used for, fun \\
                        & vacation, used for, relax out of work and school \\
                        & family, is capable of, plan to go on vacation \\
                        & X takes a family trip, because X wants, to go on vacation \\
                        & X visits the city, X is seen as, traveling \\
                        & dad, can be characterized by being, one of human's parents \\
\midrule
\multirow{14}*{Sample}  & baseball cap, used for, protect the head \\
\multirow{14}*{-\comet{}} & baseball cap, used for, protect your head while playing baseball \\
                        & baseball cap, used for, wearing over head \\
                        & X ops for baseball, but before X needs, to find a baseball \\
                        & X's favorite baseball, because X wants to, enjoy the sport \\
                        & baseball, used for, sport as a mascot \\
                        & vacation, used for, have fun on \\
                        & X chases the wind, because X wants, to walk around \\
                        & port, used for, get vacation \\
                        & field, used for, playing baseball \\
                        & cap, used for, keep their head up \\
                        & cap, used for, protection from wind \\
                        & cap, used for, protect head while traveling \\
                        & cap, used for, wear around head \\
                        & jersey, used for, wear while playing \\
\midrule
\multirow{14}*{Beam}    & baseball cap, used for, wear while playing baseball \\
\multirow{14}*{-\comet{}} & baseball cap, used for, wear on their head \\
                        & baseball cap, used for, wear on the head \\
                        & baseball, used for, playing baseball with friends \\
                        & baseball, used for, playing baseball with family \\
                        & sport cap, used for, wear while playing \\
                        & \textbf{Chicago, can be characterized by having, many streets} \\
                        & \textbf{Chicago, can be characterized by having, many cities} \\
                        & \textbf{Chicago, can be characterized by having, many neighborhoods} \\
                        & cap, used for, wear on head while playing baseball \\
                        & cap, used for, wear to the game with \\
                        & cap, used for, protect head from wind \\
                        & cap, used for, protect head from wind blows \\
                        & \textbf{cap, used for, keep the cap on} \\
                        & \textbf{cap, used for, keep the cap clean} \\
\midrule
\multirow{12}*{\diffucomet{}} & baseball cap, used for, to put on \\
\multirow{12}*{-Fact}   & \textbf{baseball cap, used for, to keep baseball cap on head} \\
                        & baseball cap, used for, wear \\
                        & baseball cap, used for, to play baseball with \\
                        & city, used for, live in \\
                        & vacation, used for, relax \\
                        & X takes a family trip, but before X needs, to spend time with family \\
                        & \textbf{X takes a family trip, because X wants, to enjoy family time} \\
                        & \textbf{X goes on vacation, because X wants, to spend time with family} \\
                        & \textbf{X is on vacation, because X wants, to spend time with family} \\
                        & dad, can be characterized by being, one of human's parents \\
                        & dad's car, used for, to be safe \\
                        & safe, used for, safe to wear \\
\midrule
\multirow{11}*{\diffucomet{}}  & cap, used for, wear on head \\
\multirow{11}*{-Entity} & cap, used for, wear on the head \\
                        & baseball cap, used for, look professional \\
                        & baseball cap, used for, to play baseball with \\
                        & \textbf{X is wearing cap, but before X needs, have a cap} \\
                        & \textbf{X is wearing cap, but before X needs, put on a cap} \\
                        & go on vacation, includes the action, take family to beach \\
                        & \textbf{go on vacation, includes the action, go somewhere nice} \\
                        & vacation, used for, enjoy your time off \\
                        & \textbf{X goes on vacation, because X wants, to spend time with family} \\
                        & dad, can be characterized by being, one of human's parents \\
                        & safe, used for, keeping things safe \\
\bottomrule
\end{tabular}
}
\caption{Examples of contextual knowledge generation. Novel and contextually relevant facts are in bold. Model notations are same as Table~\ref{tab:main_results_roc}.}
\label{tab:examples_full}
\end{table}

\begin{table}[t]
\centering
\resizebox{1.0\columnwidth}{!}{
\smallskip\begin{tabular}{lccc}
\hline
\textbf{Model} & \textbf{Physical} & \textbf{Event} & \textbf{Social} \\
\hline
Sampling-\comet{} & 46.17 & 4.01 & 49.82 \\
Beam-\comet{} & 60.72 & 2.36 & 36.92 \\
\midrule
\diffucomet-Fact & 41.00 & 4.66 & 54.34 \\
\diffucomet-Entity & 35.75 & 4.53 & 59.72 \\
\midrule
Gold & 43.54 & 7.32 & 49.14 \\
\hline
\end{tabular}
}
\caption{Proportion (\%) of different types of knowledge generation on the ROCStories portion of \comfact{}. ``Physical'', ``Event'' and ``Social'' denote facts with relation types belonging to physical-entity, event and social-interaction, respectively, as shown in Table~\ref{tab:relation_atomic}. Model notations are same as Table~\ref{tab:main_results_roc}.}
\label{tab:knowledge_type}
\end{table}

\begin{table*}[t]
\centering
\resizebox{1.0\textwidth}{!}{
\smallskip\begin{tabular}{clccccccccccc}
\hline
\multirow{2}*{\textbf{Backbone}} & \multirow{2}*{\textbf{Model}} & \multirow{2}*{\textbf{\# Facts}}  & \multicolumn{4}{c}{\textbf{Clustering \wrt{} Word-Level Edit Distance}} & \multicolumn{4}{c}{\textbf{Clustering \wrt{} Embedding Euclidean Distance}}\\
            \cmidrule(lr){4-7} \cmidrule(lr){8-11}
 &  &  & \textbf{\# Clusters} & \textbf{Relevance} & \textbf{Alignment} & \textbf{RA-F1} & \textbf{\# Clusters} & \textbf{Relevance} & \textbf{Alignment} & \textbf{RA-F1} \\
\hline
\multirow{4}*{BART} & Greedy  & 1.69 & 0.88 & 83.16 & 50.26 & 62.65 & 0.88 & 83.16 & 71.71 & 77.01 \\
\multirow{4}*{(large)} & Sampling-5   & 5.00 & 2.09 & 81.10 & 71.25 & 75.86 & 1.73 & 80.89 & 83.93 & 82.38 \\
        & Beam-5      & 5.00 & 2.12 & 82.70 & 72.69 & 77.37 & 1.64 & 82.50 & 85.78 & 84.11 \\
        & \diffucomet-Fact   & 2.56 & 1.69 & 84.39 & 74.12 & 78.92 & 1.51 & 84.38 & 86.18 & 85.27 \\
        & \diffucomet-Entity & 2.71 & 1.82 & \textbf{87.86} & \textbf{78.46} & \textbf{82.89} & 1.57 & \textbf{87.76} & \textbf{88.59} & \textbf{88.17} \\
\midrule
\multirow{2}*{\comet{}-} & Greedy  & 1.61 & 0.96 & 83.33 & 54.34 & 65.78 & 0.95 & 83.33 & 77.23 & 80.16 \\
\multirow{2}*{BART} & Sampling-5   & 5.00 & 2.09 & 80.89 & 72.77 & 76.62 & \textbf{1.76} & 80.72 & 84.21 & 82.43 \\
           & Beam-5      & 5.00 & \textbf{2.15} & 82.11 & 72.94 & 77.25 & 1.70 & 81.94 & 85.94 & 83.89 \\
\midrule
T5 (large) & Grapher & 2.10 & 1.39 & 83.48 & 70.66 & 76.54 & 1.29 & 83.46 & 82.21 & 82.83 \\
\midrule
- & Gold          & 3.22 & 2.27 & 96.43 & - & - & 1.91 & 96.43 & - & - \\
\hline
\end{tabular}
}
\caption{Clustering-based evaluation results on the \textbf{WebNLG+ 2020} benchmark. Notations are same as Table~\ref{tab:results_cluster_full_roc}.}
\label{tab:results_cluster_full_webnlg}
\end{table*}

\begin{table}[t]
\centering
\resizebox{1.0\columnwidth}{!}{
\smallskip\begin{tabular}{@{~}clc@{~~~}c@{~~~}c@{~~~}c@{~}}
\hline
\textbf{Backbone} & \textbf{Model} & \textbf{Distinct-4}  & \textbf{BLEU} & \textbf{METEOR} & \textbf{ROUGE-L} \\
\hline
\multirow{4}*{BART} & Greedy         & 87.29 & 81.12 & 84.57 & 84.92 \\
\multirow{4}*{(large)} & Sampling-5  & 48.24 & 74.22 & 81.71 & 81.19 \\
        & Beam-5       & 45.58 & 75.01 & 81.78 & 80.51 \\
        & \diffucomet-Fact   & 81.02 & 80.43 & 83.23 & 84.30 \\
        & \diffucomet-Entity & 82.20 & \textbf{83.04} & \textbf{89.88} & \textbf{89.72} \\
\midrule
\multirow{2}*{\comet{}-}     & Greedy    & \textbf{93.17} & 81.43 & 84.95 & 85.34 \\
\multirow{2}*{BART} & Sampling-5  & 47.36 & 75.44 & 81.84 & 81.85 \\
       & Beam-5       & 46.17 & 73.56 & 80.93 & 79.46 \\
\midrule
T5 (large) & Grapher & 89.95 & 76.17 & 79.61 & 80.89 \\
\midrule
  -  &      Gold     & 82.05 & - & - & - \\
\hline
\end{tabular}
}
\caption{Evaluation results of natural language generation metrics on the \textbf{WebNLG+ 2020} benchmark. Notations are same as Table~\ref{tab:results_cluster_full_roc}.}
\label{tab:results_nlg_full_webnlg}
\end{table}

\begin{table*}[t]
\centering
\resizebox{1.0\textwidth}{!}{
\smallskip\begin{tabular}{clccccccccc}
\hline
\multirow{2}*{\textbf{Backbone}} & \multirow{2}*{\textbf{Model}} & \multicolumn{3}{c}{\textbf{Exact Match}} & \multicolumn{3}{c}{\textbf{Partial Match}} & \multicolumn{3}{c}{\textbf{Strict Match}}\\
            \cmidrule(lr){3-5} \cmidrule(lr){6-8} \cmidrule(lr){9-11}
 &  & \textbf{Web-Prec.} & \textbf{Web-Rec.}  & \textbf{Web-F1} & \textbf{Web-Prec.} & \textbf{Web-Rec.}  & \textbf{Web-F1} & \textbf{Web-Prec.} & \textbf{Web-Rec.}  & \textbf{Web-F1} \\
\hline
\multirow{4}*{BART} & Greedy  & 50.42 & 52.79 & 51.51 & 53.76 & 56.84 & 55.20 & 50.14 & 52.53 & 51.25 \\
\multirow{4}*{(large)} & Sampling-5   & 73.65 & 76.73 & 75.11 & 79.57 & 83.89 & 81.66 & 72.37 & 75.45 & 73.83 \\
        & Beam-5      & 75.32 & 78.39 & 76.76 & 81.32 & 85.72 & 83.38 & 73.36 & 76.27 & 74.75 \\
        & \diffucomet-Fact   & \underline{76.59} & 78.35 & \underline{77.47} & 79.17 & 81.52 & 80.35 & \underline{76.30} & \underline{78.07} & \underline{77.19} \\
        & \diffucomet-Entity & \textbf{80.80} & \textbf{82.97} & \textbf{81.84} & \textbf{83.72} & \textbf{86.48} & \textbf{85.07} & \textbf{80.68} & \textbf{82.89} & \textbf{81.74} \\
\midrule
\multirow{2}*{\comet{}-} & Greedy  & 52.55 & 54.82 & 53.62 & 55.99 & 58.95 & 57.39 & 52.30 & 54.59 & 53.37 \\
\multirow{2}*{BART} & Sampling-5   & 74.96 & 77.87 & 76.33 & 80.31 & 84.41 & 82.18 & 73.77 & 76.67 & 75.15 \\
           & Beam-5      & 75.95 & \underline{78.88} & 77.03 & \underline{81.66} & \underline{85.84} & \underline{83.15} & 73.80 & 76.61 & 74.85 \\
\midrule
T5 (large) & Grapher & 71.50 & 73.30 & 72.20 & 74.10 & 76.50 & 75.00 & 71.20 & 73.00 & 71.90 \\
\hline
\end{tabular}
}
\caption{Evaluation results on official metrics provided by the \textbf{WebNLG+ 2020} benchmark challenge. We present the results of Grapher as reported in its paper. Notations are same as Table~\ref{tab:results_cluster_full_roc}.}
\label{tab:results_official_full_webnlg}
\end{table*}